\documentclass[a4paper]{article}

\usepackage[top=2.5cm, bottom=2.5cm, left=2.5cm, right=2.5cm]{geometry}

\usepackage{graphicx}

\usepackage{amsmath} 
\usepackage{footnote}
\usepackage{url}
\usepackage{amssymb}

\usepackage{natbib} 

\begin{document}

\title{Automatic text summarization: What has been done and what has to be done}

\author{Abdelkrime Aries \and Djamel eddine Zegour \and Walid Khaled Hidouci}
\date{\textit{Ecole nationale Sup\'erieure d'Informatique (ESI, ex. INI), Algiers, Algeria} \\[.2cm]
	Emails: \{ab\_aries, d\_zegour, w\_hidouci\}@esi.dz}

\maketitle

\begin{abstract}
Summaries are important when it comes to process huge amounts of information. 
Their most important benefit is saving time, which we do not have much nowadays.
Therefore, a summary must be short, representative and readable.
Generating summaries automatically can be beneficial for humans, since it can save time and help selecting relevant documents.
Automatic summarization and, in particular, Automatic text summarization (ATS) is not a new research field; It was known since the 50s. 
Since then, researchers have been active to find the perfect summarization method.
In this article, we will discuss different works in automatic summarization, especially the recent ones.
We will present some problems and limits which prevent works to move forward.
Most of these challenges are much more related to the nature of processed languages. 
These challenges are interesting for academics and developers, as a path to follow in this field.

\end{abstract}

\noindent
{\it Keywords:} 
Text summarization,
 summary evaluation,
 summarization methods

\section{Introduction}

A summary, as defined in TheFreeDictionary\footnote{TheFreeDictionary: \url{http://www.thefreedictionary.com/outline} [07 February 2017]}, is ``\textit{a brief statement that presents the main points in a concise form}".
One of its benefits is identifying the content of a given document, which will help readers peaking the right documents to their interests.
According to \citet{75-borko-bernier}, a summary has many benefits such as saving time, easing selection and search, improving indexing efficiency, etc. 
Summaries of the same document can be different from person to person. 
This can occur due to different points of interest or due to each individual's understanding.

Automatic text summarization is not a new field of research. 
Since the work of \citet{58-luhn}, many works have been conducted to improve this field of research. 
Even if it was an old field of research, ATS still gains a lot of attention from research communities.
This field is motivated by the growth of electronic data turning the selection of relevant information into a more difficult task.
Information can be found in many sources, such as electronic newspapers, with many variations.
The redundant information makes their selection and processing very difficult.

Our goal is to afford a concise report about different research works in ATS. 
This can help academics and professionals having an overview on what has been already done in this wide field. 
Also, we indicate some challenges to lead future research.
Most of them are considered as related topics to this field, since they are associated with natural language processing (NLP) domain.
Though these challenges are well known, they are still holding this field from going forward.

The remainder of this article is organized as follows. 
Section \ref{sec:summeth} covers the different methods in ATS classified using criteria of input document, purpose and output document. 
Section \ref{sec:app} is an enumeration of some major approaches for ATS.
Section \ref{sec:eval} presents the evaluation methods, workshops and campaigns. 
These three sections are intended to present what has been done in ATS.
Section \ref{sec:challenges} addresses some challenges a researcher may find in this field.
This late section is intended to present what has to be done in order to advance this field forward.
Finally, section \ref{sec:conclusion} sums up what has been discussed in all of these sections.

\section{Summarization methods}
\label{sec:summeth}

\begin{figure}[ht]
\begin{center}
	\includegraphics{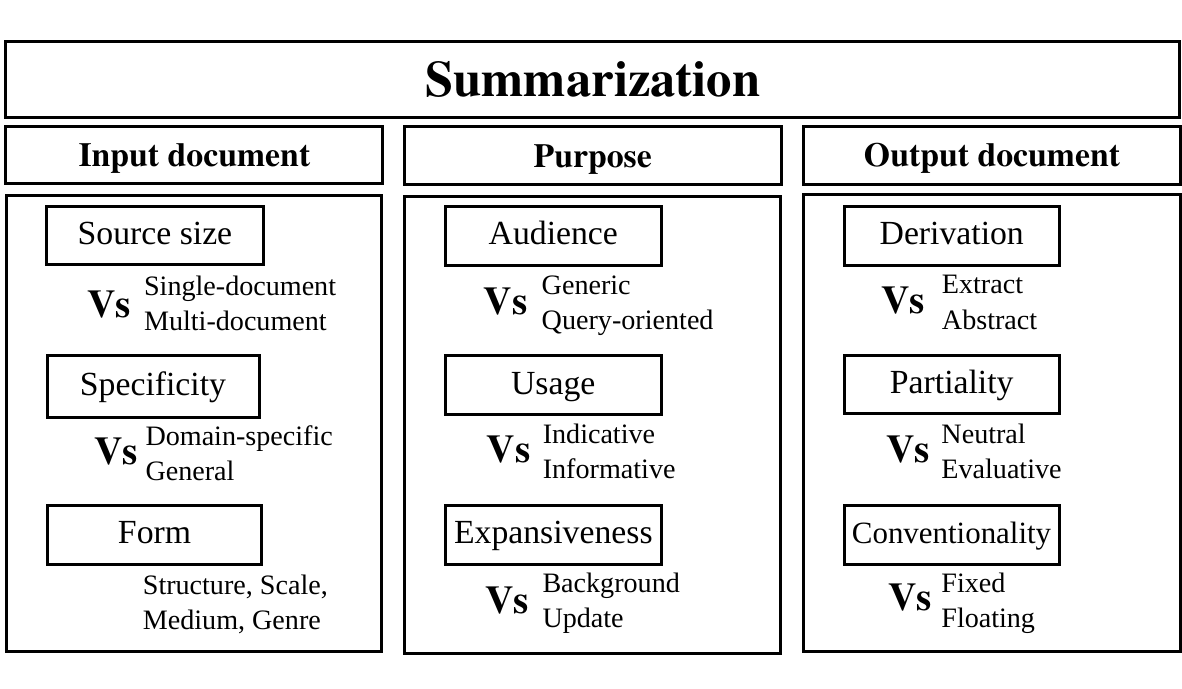} 
	\caption{Classification of summarization systems using different criteria.}
	\label{fig:summary-classif}
\end{center}
\end{figure}

Different summarization methods have been proposed since the first work of Luhn. 
Each method is intended to solve a problem in a certain context. 
A summarization system can be classified using many criteria; It can belong to many classes at once. 
These criteria are regrouped into three categories: input document criteria, purpose criteria and output document criteria \citep{98-hovy-lin,99-sparckjones}. 
Based on the criteria afforded by \citet{98-hovy-lin} and some other changes to fit current systems, the different classifications can be illustrated as in Figure \ref{fig:summary-classif}.

\subsection{Input document}

Examining the input document, we can classify a summarization system using three criteria: source size, specificity and form. 
The source size refers to how many documents a system can have as input. 
The specificity refers to the domains which this system can handle: is it designed just for a specific domain? or is it a general purpose system? 
The form of the input document(s) specifies if they are structured or not, have low scale (tweets) or large one (novel), are textual or multimedia documents, and if they are of the same genre or not. 

\subsubsection{Source size}

A summarization system can have one input document or many. 
Mono-document summarizers process just one input document even if the summarizer used some other resources for learning. 
This kind of summarization includes the first works of ATS \citep{58-luhn,58-baxendale,69-edmundson}.

In the other hand, Multi-document summarizers take as input many documents of the same topic (for example, news about a car crush from many newspapers). 
The earliest work we can find in multi-document summarization is the work of \citet{95-mckeown-radev}. 
The benefit of this type of summarization is to diversify information sources. 
But, the problem is the redundancy; Documents talking about the same topic have much information in common.
This classification (Mono vs. Multi- document summarization) is very popular among researchers in the field of ATS.

\subsubsection{Specificity}

An automatic summary can be generated from domain-specific documents or from general domains.
When we want to summarize documents of the same domain, it is more appropriate to use a summarization system specific to this domain. 
This can reduce terms ambiguities, the usage of grammar and formating schemes. 
Some works use domain specific concepts derived from some resources to enhance relevancy detection.
In biomedical text summarization, the work of \citet{07-reeve-al} uses two methods: one to identify salient sentences and the other to remove redundant information. 
The two methods are based on concepts extracted from biomedical resources, and are both used to generate the final summary.
In medical text summarization, \citet{09-sarkar} uses domain-specific cue phrases combined with other features to decide sentence relevancy. 
LetSumm system \citep{04-farzindar-lapalme} is also an example of domain-specific summarization systems designed for juridic texts.

\subsubsection{Form}

Input documents can have many forms based on their structures, scales, mediums and genres. 
The structure refers to the explicit organization found in the document (eg. sections in papers: introduction, related works, method, experiments and conclusion).
The structure can be used to increase summarization performance where every section can be treated differently. 
For instance, LetSumm system \citep{04-farzindar-lapalme} uses thematic structures in legal judgments (decision data, Introduction, Context, juridical analysis and Conclusion) to generate summaries. 
For each thematic, some linguistic markers are affected to be used as cue phrases \citep{81-paice}. 
The resulted summary is the combination of those thematics summaries with a percentage of contribution.
\citet{07-pembe-gungor} use HTML documents as input document. 
They score the sentences, extract the first ones and reorder them using their initial sections (the sections are detected using HTML tags).

The scale is another aspect which can affect the summarizing method; An input document can be a paragraph, an article, a book, etc. 
Many known summarization systems use term frequency as relevancy measure, therefore the input document must be large (important terms are repeated a lot).
In cases of summarizing low scale documents such as tweets and micro-blogs, we can't use usual techniques.
The work of \citet{10-sharifi-al} revolves around the summarization of tweets.
Their algorithm takes as input a trending phrase or a phrase specified by user.
This phrase is used to collect a large number of tweets which are served to generate the summary.
In this case, the summary is the most commonly used phrase covering the topic phrase. 
In \citep{12-duan-al}, given a hashtag as a topic, we can select a set of tweets.

The medium is the support which carries the information.
Most summarization systems are based on textual support, while there are some researches on summarizing images, videos and audios.
The work of \citet{14-gunhee-al} addresses the problem of jointly summarizing large-scale Flickr images and YouTube user videos. 
The summarization of videos uses similarity with images to delete redundant information. 
In the other hand, summarizing photos consists of creating a photo storyline with the assistance of videos.
Lately, researchers show some interest in multimedia summarization: interactive football summarization \citep{09-moon}, summarizing important events in a football video \citep{12-zawbaa-al}, video summarization using web images \citep{13-khosla-al}, video summarization using a given category \citep{14-potapov-al}, video summarization by learning \citep{15-gygli-al}.

Summarization systems can be classified using the genre of their input documents: news, interviews, reviews, novels, etc.
There are some research works which adjust their scoring methods according to the document's genre. 
In \citep{07-goldstein-al}, the authors propose a machine learning based method which detects the genre of the input document and estimates the adequate features.
A similar work can be found in \citep{10-yatsko-al}, where the summary is generated based on the genre.
We can find some systems designed for a specific genre, such as reviews \citep{09-zhan-al}.

\subsection{Purpose}


\subsubsection{Audience}

Sometimes, the user needs a summary which focuses on some aspects rather than the main idea of the input document. 
A summarization system which takes in consideration the user's preferences is called query-oriented, while the one which does not is known as generic system. 
A generic summarization system tries to preserve important information out of an input document. 
So, we can say it tries to focus on the topic presented by the author and present what the input document is all about. 
This type of systems exists widely, as an example: the systems in MultiLing'15 \citep{15-vanetik-litvak,15-aries-al,15-thomas-al,15-vicente-al}.

Query-oriented systems take in consideration the users preferences.
For instance, if a user wants a summary focusing on someone in a story, it must contain events around this person without loosing the main interest of the story. 
\citet{10-bhaskar-bandyopadhyay} describe a method to summarize multiple documents based on user queries.
The correlation measure between sentences from each document are calculated to create clusters of similar sentences.
Then, the sentences are scored according to the query.
This score is accumulated to the cluster score to extract the highest scored sentences.
Texts are not the only thing to summarize, concepts can be summarized too.
In \citep{15-cebiric-al}, the authors propose a method for query-oriented summarization using RDF\footnote{RDF: Resource Description Framework} graphs.
In their method, they try to summarize an RDF graph to get another one which focuses on a given query. 
Recent techniques such as deep learning have been used to generate query-driven summaries \citep{15-zhong-al}.

\subsubsection{Usage}

The usage of a summary searches whether it is meant to help users decide their interests or as a representative replacement for the original document(s).
Informative summaries contain the essential information of the original document(s); After reading them, you can tell what are the main ideas. 
This type of summaries can be found in journals, thesis, research articles, etc. 
In a research article, the author must present the essential ideas and findings using an abstract which is considered as an informative summary.
The majority of systems and methods focus on this type of summarization, hence we can find examples of this type anywhere.

\begin{figure}[ht]
	\begin{center}
		\includegraphics{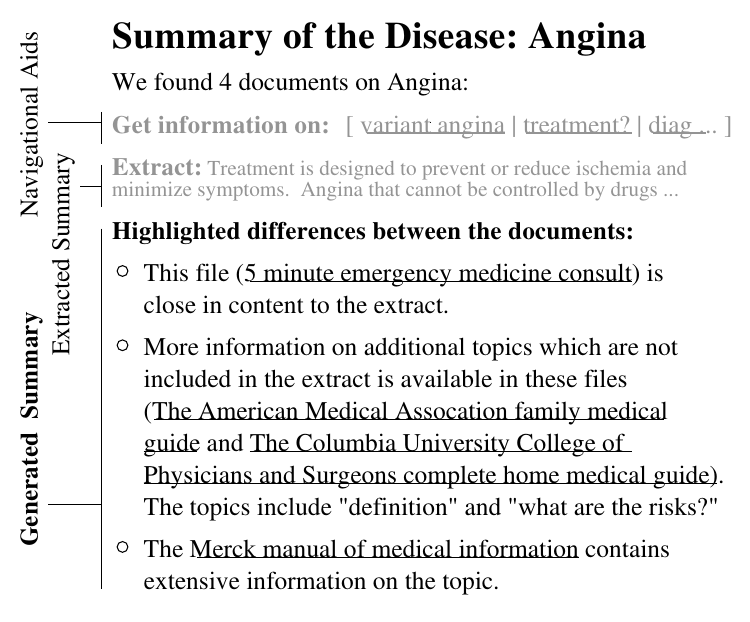} 
		\caption{An example of CENTRIFUSER summarizer output \citep{01-kan-al}.}
		\label{fig:kan-al-exp}
	\end{center}
\end{figure}

Indicative summaries does not contain informative content; they only contain a global description of the original document. 
That includes the purpose, scope, and research methodology. 
This can be helpful to decide whether or not to consult the original document. 
The abstracts on the title page's verso of books and reports are an example of indicative summaries. 
Works have been conducted to investigate such type of summaries.
\citet{01-kan-al} propose an indicative multi-document summarization system (CENTRIFUSER) based on the problem of content planning. 
Figure \ref{fig:kan-al-exp} represents an example of a CENTRIFUSER summary on the health-care topic of ``Angina" where the generated indicative summary is in the bottom half using the difference in topic distribution of documents to regroup them.
The method uses an IR system to search in a collection of documents, which gives a selected set of documents. 
Then some features are extracted from the collection and from the individual chosen documents to be used along with the query to generate the indicative summary.
%
%

\subsubsection{Expansiveness}

A generated summary can focus on the background of the original document, or affords the news compared to some past documents.
This property is referred to as expansiveness. 
A background summary, according to \citet{01-mani}, assumes that the reader has poor prior knowledge of general setting of the input text(s), and hence includes explanatory material, such as circumstances of place, time, and actors.
In the other hand, just-the-news summary is containing just novel or principal themes, assuming that the reader
knows enough background to interpret them in context. 
Nowadays, a lot of summarizers tend to afford the most relevant information found in input texts without verifying if the resulted summary incorporates some explanatory materials. 
Following the previous two definitions, these summarizers are neither background nor just-the-news. 
That is, a summarizer of this kind can generate both background summaries and just-the-news summaries. 
So, classifying systems based on whether they are designed to incorporate news is more appropriate.

Lets, first, redefine what just-the-news summaries and call them update summaries, since this term is more used. 
Recently, the emerging interest in systems that automatically monitor streams of social media posts such as Twitter or news to keep users up to date on topics they care about, led to the appearance of update summaries.
The idea is to generate summaries from some recent documents which do not contain information from previous documents. 
It means, the system must have prior knowledge of what have been seen before. 
Update summaries were promoted in DUC 2007's ``update task" where the participants must generate multi-document update summaries from some chronologically ordered sets of news documents (see section \ref{sssec:eval-duc} for more detail).
More recent evaluation campaigns for update summaries are: TREC's Temporal Summarization and Real-time summarization tasks.

We can, then, define a background summarization system as a system which generates summaries based on the input document(s) content and without excluding information from prior documents on the same subject. 

\subsection{Output document}

Three criteria of the summary can be used to classify a summarization system: derivation, partiality and format.
The derivation is the way used to produce a summary from the original document, either by extracting relevant units or by creating a new text.
Partiality is how a summary handles the opinions found in the original document, which can be either neutral or evaluative.
As for format, the summary format can be fix or floating.
 
\subsubsection{Derivation}

It refers to the way used to obtain a summary. 
It can be by extracting pertinent units or by understanding and generating a new summary.

Extractive summaries are produced by, as the name says, extracting units from the original documents. 
Usually, these units are sentences because it is easy to keep the correctness of the grammatical structure.
The first researches in the field of ATS are extractive; they use some features to estimate the pertinence of a sentence.
Some of these features are: term frequency \citep{58-luhn}, position in the text \citep{58-baxendale,69-edmundson} and keywords \citep{69-edmundson}. 
Till nowadays, this type of summarization is the most used \citep{15-vanetik-litvak,15-aries-al,15-vicente-al,15-thomas-al}.
What makes extractive methods so famous is their simplicity compared to the abstractive ones. 

Abstraction, in the other hand, is the generation of new text based on the input document(s).
Abstractive systems are difficult to be designed due to their heavy dependence on linguistic techniques.
Specifying the domain of the system can simplify the creation of this type of systems \citep{93-mitkov}.

\subsubsection{Partiality}

Partiality, by definition, is the bias in favor of one thing over another. 
following partiality, a summarization system can be neutral or evaluative.
A neutral system produces summaries which reflect the content of the input document(s) without judgment or evaluation. 
They are not designed to, specifically, include opinion into the summary even if the input document contains judgments.
Most works fall into this class \citep{58-luhn,58-baxendale,69-edmundson,15-vanetik-litvak,15-aries-al,15-vicente-al,15-thomas-al,15-zhong-al}.

In contrast, an evaluative system includes automatic judgments either implicitly or explicitly.
An explicit judgment can be seen as some statements of opinion included.
The implicit one uses bias to include some material and omit another.
A lot of examples can be afforded for this type of summarization, especially with the growth of interest towards users opinions. 
For instance, the work of \citet{16-othman-al} is based on summarizing customer opinions through Twitter. 
Given a conversation of tweets, the method tries to effectively extract the different product features as well as the polarity of the conversation messages.
The year 2008 knew the debut of a new task within TAC conference called ``Opinion Summarization task" which was meant to generate summaries from answers on opinion questions (see Section \ref{sssec:eval-duc}) 

\subsubsection{Format}

Each system has a format used to present the resulted summaries. 
Some systems use a fixed format, while others present the summaries based on user preferences or based on their goals.
Once again, most systems generate a fixed format mostly by joining sentences together.
Most of them are research systems which focus on the informational part of the summary rather than its format (presentation to the user).
Floating-situation summarization systems try to display the content of summaries using variable settings to a variety of readers for diverse purposes. 
The ONTOSUM system of \citet{05-bontcheva} is one of these systems, which uses device profile (e.g., mobile phone, Web browser) to adjust the summary formatting and length.

\section{Approaches}
\label{sec:app}

In this section, we will discuss different approaches of ATS. 
According to \citet{12-nenkova-mckeown}, topic representation approach contains topic words, frequency driven, latent semantic analysis, etc. 
We are, rather, interested in the nature of used resources, are they dependent on a specific language? or domain? do the methods need a lot of resources?
This is why we follow their other taxonomy \citep{11-nenkova-mckeown} called ``Semantics and discourse" which we present as ``linguistic approach" since its methods as highly connected to the language being processed.
The taxonomy of \citet{12-lloret-palomar} seems to approach our vision. 
The difference is that we consider topic-based and discourse-based approaches as sub-approaches of linguistic one, since the two are based on linguistic properties of the input text.

\subsection{Statistical}

Statistical approach has been used in ATS since its first days. 
It is based on some features which are used to score the relevancy of text units (generally, sentences) to the main topic or users requests. 
These features can be combined to score the units using many aspects, and get the highest scored ones.
But, combining some features does not always improve the quality of a summary. 
As follows, we will present the most used amongst statistical features.

\subsubsection{term frequency}

It is the oldest feature \citep{58-luhn} and the most famous one. 
It supposes that a term repeated many times in a text is important.
This feature has a problem when it comes to domain-relative words; For example, in documents talking about computer science, certain words such as ``computer" will have great frequencies even if they do not represent the main topic.
To address this problem, \citet{58-luhn} uses two thresholds to ensure that the term is important yet not specific to the document's domain.
A more advanced solution is to use $ tf*idf $ defined by \citet{73-salton-yang}.
The measure $ idf $ (inverse document frequency) is calculated as in Equation~\ref{eq:idf}.
\begin{equation}
\label{eq:idf}
idf(t) = log{\frac{|D|}{|\{d /\ t \in d\}|+1}}
\end{equation}
Where: 
$ |D| $ is the number of documents in corpus $ D $;
$ |\{d /\ t \in d\}| $ is the number of documents containing $ t $.

\subsubsection{Position in the text}

\begin{figure}
\begin{center}
	\includegraphics[width=50mm]{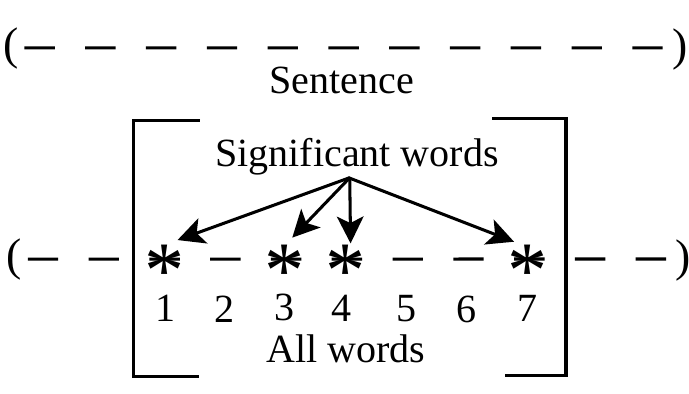} 
	\caption{Luhn score using word position \citep{58-luhn}.}
	\label{fig:luhn-score}
\end{center}
\end{figure}

The position of words \citep{58-luhn} and sentences \citep{58-baxendale,69-edmundson} in the text has good potential to capture their importance. 
In \citep{58-luhn}, the position of words in a sentence is used to create some sort of groups, where each one is the set of significant words separated by at most 5 to 6 non-significant words. 
Then, the group which has the most significant words is used to score the sentence, as shown in Figure \ref{fig:luhn-score}.
The position of sentences in the text is used as indicator of its importance: the first and the last sentences tend to be more informative than the others \citep{69-edmundson}.
For instance, in scientific articles, sentences from the  introduction and the conclusion contain more information about the subject than the other sentences.
In \citep{58-baxendale}, it is established that the first and the last sentences in paragraphs are more important. 

Position feature can be used differently to score the sentences.
\citet{04-nobata-sekine} define three methods to score a sentence based on its position in the input document: only the first sentences less than a defined position are important, a sentence's importance is inversely proportional to its position, and the first and the last sentences are more important.
According to the authors, the second method gives the best results among the three.
\citet{09-abdelfattah-ren} use the position of sentences in paragraphs instead of the whole text. 
They suppose that the 5 first sentences in a paragraph are the most important, and thus the others should have a score of zero.

Like Luhn, \citet{10-ouyang-al} use word position based on the hypothesis that a word is more informative if it appears earlier in the text.
Therefore, the position of a word can be calculated according to its other occurrences in the whole text and not just according to other words in the sentence.
For that, four different functions are defined: Direct proportion (DP), Inverse proportion (IP), Geometric sequence (GS) and Binary function (BF). 
DP attributes a score of $ 1 $ of the first appearance and $ 1/n $ to the last one, where $ n $ is the count of words in the sentence. 
IP is calculated as $ \frac{1}{i} $, where the degree decreases quickly at smaller positions. This gives advantage to leading sentences. 
GS function scores an appearance of a word as the sum of the scores of all its following appearances as $ (1/2)^{i-1} $.
BF affords more importance to the first appearance of a word, and the others an equally less importance.
Therefore the first one will get a score of $ 1 $ and the others a score of a given $ \lambda \ll 1 $.
The final score is calculated as shown in Equation \ref{eq:ouyang-pos} 
\begin{equation}
\label{eq:ouyang-pos}
Score(s) = \sum\limits_{w_i \in s} \frac{\log(freq(w_i)) * pos(w_i)}{|s|} 
\end{equation}
Where $ pos(w_i) $ is one of the four functions shown previously; $ freq(w_i) $ is the frequency of the word $ w_i $; $ |s| $ is the length of the sentence.

\subsubsection{Title and subtitle words}

``The title carries the topic of the document", this hypothesis was first introduced by \citet{69-edmundson}.
When we divide a document into sections and subsections, we choose representative titles for each. 
So, any sentence containing some words of a title is considered as important.
This feature can be seen as if the title was a request \citep{88-salton-buckley}. 

\citet{01-ishikawa-al} fuse this feature with term frequency, in a way that the frequencies of the title words are weighted more than regular words.
In Equation \ref{eq:ishikawa-head}, a sentence $ s_i $ is scored by the frequencies of its words $ w $; if the word belongs to the title, its frequency is multiplied by a number $ A > 1 $ (the authors used $ A = 3 $).
\begin{equation}
\label{eq:ishikawa-head}
Score_{title}(s_i) = \sum_{\{w\} \in s_i}{\alpha(w) * tf(w)} 
\text{ where }
\alpha(w) = \left\lbrace 
\begin{array}{ll}
A > 1 & \text{if } w \in title \\
1 & otherwise \\
\end{array} 
\right. 
\end{equation}
To score sentences based on title's words, \citet{04-nobata-sekine} propose two methods.
According to the authors, the second method, shown in Equation \ref{eq:nobta-head-2}, gives better results.
\begin{equation}
\label{eq:nobta-head-2}
Score_{title}(s_i) = \frac{\sum_{e \in T \bigcap s_i}{\frac{tf(e)}{tf(e)+1}}}
{\sum_{e \in T}{\frac{tf(e)}{tf(e)+1}}}
\end{equation}
Where, $ e $ are named entities, and $ tf $ is term frequency.

\subsubsection{Sentence length}

This feature was used to penalize too short sentences \citep{95-kupiec-al}.
Given a threshold, for example 5 words, the feature is true if it exceeds this threshold and false otherwise.
Sentences which are very small, in number of words, are unlikely to be important so it is better to omit them.
A more complex formula to score a sentence is expressed in the two methods proposed by \citet{04-nobata-sekine}.
The first method scores a sentence based on its length and a predefined maximum value $ L_{max} $.
The second, which gives better results, affords a negative score to penalize sentences shorter than a predefined minimum value $ L_{min}$.
A more recent formula proposed by \citet{09-abdelfattah-ren} is indicated in Equation \ref{eq:abdelfattah-len}.
It scores a sentence $ s_i $ using its words number $ |s_i| $, the document's words number $ |d| $ and the number of sentences in this document $ |\{s: s \in d\}| $.
\begin{equation}
\label{eq:abdelfattah-len}
Score_{length}(s_i) = \frac{|s_i| * |\{s: s \in d\}|}{|d|}
\end{equation}

\subsubsection{Centroid}

A centroid, as defined by \citet{04-radev-al}, is ``\textit{a set of words that are statistically important to a cluster of documents}". 
Since it is the most important in the cluster, the documents and sentences containing it are also important.
One famous summarization system using clusters centroid to extract summaries is MEAD \citep{00-radev-al,04-radev-al}.
MEAD is a multi-document summarizer, where similar documents to the centroid are clustered together.
Then, a set of parameters (Centroid value, Positional value, and First-sentence overlap) are used to rank each sentence from the resulting cluster.
The centroid value $ C_i $ for a sentence $ S_i $ is computed as $ C_i = \sum_{w \in S_i}  C_{w,i} $.
Where, $ C_{w,i} $ is the centroid value of a word $ w $ in sentence $ S_i $.
The positional value $ P_i $ for a sentence $ S_i $ is computed according to its position in the document with $ n $ sentences as follows: $ P_i = \frac{n - i + 1}{n} * C_{max} $ where $ C_{max} $ is the maximum centroid score in that document.
The First-sentence overlap value $ F_i $ of a sentence $ S_i $ is given as $ F_i = \overrightarrow{S_1} \overrightarrow{S_i} $, where $ S_1 $ is the first sentence.
These three scores are combined into one score along with a redundancy score to extract the most scored sentences.

\subsubsection{Frequent itemsets}

Frequent itemsets are common sets of items which have at least a minimum amount of times. 
It was, first, proposed by \citep{94-agrawal-srikant} to solve the problem of discovering association rules
between items in a large database of sales transactions.
In ATS, the itemsets are considered as sets of terms extracted from sentences, where those which co-occur in many sentences are considered as frequent itemsets.

\citet{12-baralis-al} apply an entropy-based strategy to generate compact itemset-based models, where each document sentence is seen as a separate transaction. 
They formalize the problem of selecting sentences as a set covering optimization problem which is solved by a greedy strategy where sentences covering the maximum number of itemsets.
This method is ameliorated in MWI-Sum system \citep{15-baralis-al} by replacing traditional itemsets with weighted ones in order to increase item relevance in the mining process. 
Also, term frequency-document frequency ($ tf-df $) is used as relevance score. 

This approach is used in \citep{17-litvak-vanetik} with the minimum description length (MDL) principle that employs
Krimp compression algorithm \citep{11-vreeken-al} for query-based ATS.
The key idea is to use the query to select related frequent itemsets (word sets).

\subsubsection{Latent semantics}

Latent semantic analysis (LSA) seeks to analyze relationships between a set of documents and the terms they contain.
It was used in \citep{04-steinberger-jezek} for text summarization. 
The algorithm starts by creating a matrix $ A $ of $ m $ rows representing the document terms, and 
$ n $ columns representing the sentences where $ a_{i, j} \in A $ represents the frequency of the term $ i $ in the sentence $ j $. 
Then, the singular value decomposition (SVD) of the matrix $ A $ is represented as in equation \ref{eq:svd}.
\begin{equation}
\label{eq:svd}
A = U \varSigma V^T
\end{equation}
Where:
\begin{itemize}
	\item $ U = [u_{i,j}] $ is an $ m \times n $ column-orthonormal matrix whose columns are called left singular vectors.
	\item $ \varSigma = diag(\sigma 1, \sigma 2, ..., \sigma n) $ is an $ n \times n $ diagonal matrix, whose diagonal elements are non-negative singular values sorted in descending order
	\item $ V = [v_{i,j}] $ is an $ n \times n $ orthonormal matrix, whose columns are called right singular vectors.  
\end{itemize}
Then, the salience of a sentence $ k $ is given in equation \ref{eq:lsa}.
\begin{equation}
\label{eq:lsa}
s_k = \sqrt{\sum\limits_{i=1}^{n} v_{k,i}^2 . \sigma^2}
\end{equation}


\subsubsection{Combination of features}

Usually, to score a unit (sentence) not just one feature but many are used as shown in equation \ref{eq:combin-crit}.
\begin{equation}
\label{eq:combin-crit}
Score(s_i) = \sum_{f \in F}{\alpha_f * Score_f(s_i)}
\end{equation}
Where $ \alpha_f $ is the weight of a feature $ f $ in the features set $ F $.
An early work which combine many features to score sentences is the work of \citet{69-edmundson}.
The author used four features: Cue words (\textit{C: Cue}), term frequency (\textit{K: Key}), title words (\textit{T: Title}) and position (\textit{L: Location}).
He used all the combinations of these features which gives 15 different combinations, where the weights are equal to 1.
Most systems use different features combined together (linear combination, generally) to get a unique score. 
The features weights can be fixed manually or using machine learning techniques described in a later section.
Also, other methods such as optimization can be used. 
For instance, \citep{17-feigenblat-al} propose an unsupervised query-focused multi-document summarization method based on Cross-Entropy (CE) method \citep{04-rubinstein-kroese}. 
It seeks to select a subset of sentences that maximizes a given quality target function using some statistical features.

\subsection{Graphs}

Inter-cohesion between text units (sentences) is an important property; A summary which contains linked units has the chance to be more pertinent to the input document's main topic.
Graph-based approach is based on transforming documents units (sentences, paragraphs, etc.) into a graph using a similarity measure, then this structure is used to calculate the importance of each unit. 
We divide graph-based works into two categories: those using graph properties such as the number of neighbors, and those iterating over the graph and changing nodes' scores till reaching a stable graph representation.

\subsubsection{Graph properties}

In \citep{97-salton-al}, document paragraphs are used to construct a graph of similarities, where each paragraph represents a node which is connected to another when their similarity is above a given threshold.
The authors define a feature called bushiness which is the number of a node's connections. 
The most scored paragraphs in term of bushiness are extracted to form a summary.

Node's bushiness can be used to score sentences alongside other statistical features \citep{09-abdelfattah-ren}.
In a graph of similarities between sentences, the importance of a sentence is the number of arcs connected to it.
Equation \ref{eq:abdelfattah-arcs} represents the score based on number of arcs, where $ G =  \{S, A\} $ is the graph of similarities between the sentences, $ S $ is the set of sentences and $ A $ is the set of arcs.
\begin{equation}
\label{eq:abdelfattah-arcs}
Score_{\#arcs}(s_i) = |\{ s_j : a(s_i, s_j) \in A / s_j \in S, s_i \neq s_j \}|
\end{equation}

Beside Bushy Path of the Node, \citet{14-ferreira-al} define another property called ``Aggregate Similarity". 
Instead of counting the number of arcs connected to a node, their weights are summed to represent this node's importance.
In their work, the authors investigate the fusion of multiple features, either graph-based or sentence features.

\subsubsection{Iterative graph}

LexRank \citep{04-erkan-radev} and TextRank \citep{04-mihalcea-tarau} are the most popular methods using graph-based summarization approach. 
The two methods use a modified version of PageRank \citep{98-brin-page} in order to score sentences. 
LexRank method \citep{04-erkan-radev} uses cosine similarity to construct a weighted graph where the nodes with a weight (similarity) less than a given threshold are omitted. 
The continuous version follows almost the same equation as of TextRank, but it uses a $ tf-idf $ based cosine similarity instead.
%
In TextRank, an indirected graph is constructed from the input text, where each sentence represents a node, and the arc between two nodes is weighted by their similarity. 
Equation \ref{eq:textrank} scores each sentence $ i $ based on its neighbors, and it is executed recursively till reaching convergence, where $ d $ is the damping factor (usually around $ 0.85 $).
\begin{equation}
\label{eq:textrank}
WS(V_i) = ( 1 - d) + d * \sum\limits_{V_j \in In(V_i)} \frac{w_{ji}}{\sum\limits_{V_k \in Out(V_j)} w_{jk}} WS(V_j)
\end{equation}
Where:
\begin{equation}
\label{eq:textrank-sim}
w_{ij} = \frac{|\{w_k \text{ / } w_k \in S_i \text{ and } w_k \in S_j\}|}{\log(|S_i|) + \log(|S_j|)}
\end{equation}

\citet{06-li-al} use PageRank to rank document events rather than sentences, then extract those sentences containing more important events.
They define an event as either a named entity (NE: Person, Organization, Location or Date), or an event term (ET: verb or action noun).
The relevance between two events is used to weight the arcs relating them.
It is calculated as the number of association in case of a pair of ET and NE.
In case of an ET pair, it is calculated as the number of NE associated to both of them. 
Similarly, in case of an NE pair, it is the number of ET associated to both of them. 

In case of multi-document summarization where there are some documents more recent than others, the temporal information matters. 
Recent documents contain novel information in an evolving topic, therefore their sentences may be given more chance to be included into the summary. 
\citet{07-wan} proposes a method called TimedTextRank to incorporate this information into TextRank method.
The informativeness score ($ WS(V_j) $) is time-weighted, multiplying it by $ 0.5^{(y - t_j)/24} $ where $ y $ is the current time and $ t_j $ is publication time of the document containing a sentence $ j $.

TextRank and LexRank exploit sentence-to-sentence relationships to score them, under the assumption that they are indistinguishable.
But in multi-document summarization, a document may be more important than others and therefore its sentences must be favored over others.
\citet{08-wan} proposes adding a sentence-to-document relationship into the graph-based ranking process. 
In addition to documents impact on sentences, the author argues that even sentences in the same document must not be treated uniformly.
The position of a sentence and its distance to the document's centroid are two factors to be included in sentence score.

Most graph-based summarization methods \citep{04-mihalcea-tarau,04-erkan-radev,06-li-al,08-wan} are based on ranking algorithms developed for web-pages analyze, such as PageRank \citep{98-brin-page} and HITS \citep{99-kleinberg}. 
In their method called iSpreadRank, \citet{08-yeh-al} exploit activation theory \citep{68-quillian} which explains the cognitive process of human comprehension.
The idea is to construct a graph of similarities between sentences, score each sentence using some features (centroid, position, and First-sentence overlap), then spread the scores to the neighbors iteratively until reaching equilibrium.


Some works tried to introduce machine learning into graph-based ATS. 
In \citep{08-liu-al}, a prior probability is incorporated into PageRank algorithm to introduce query relevance into graph-based approach. 
The prior probability $ p(s/q) $ is estimated using Naive Bayes where the relevance of a sentence $ p(s) $ is estimated using four features (Paragraph Feature, Position in Paragraph Feature, Mood Type Feature, and Length Feature), and the relevance of query having some sentences $ p(q/s) $ is estimated using shared named entities between a query and the sentences in the training corpus.
Then, this probability is introduced to the past LexRank equation (see Equation \ref{eq:textrank}) as in Equation \ref{eq:pprsum}.

\begin{equation}
\label{eq:pprsum}
WS(V_i) = d * p(V_i/q) + (1-d) * \sum\limits_{V_j \in In(V_i)} \frac{w_{ji}}{\sum\limits_{V_k \in Out(V_j)} w_{jk}} WS(V_j)
\end{equation}
This model can select sentences with high relevance to the query without loosing the ability to select those with high information novelty.

\subsection{Linguistic}

Statistical approach uses some primary NLP techniques such as word segmentation, stop word elimination and stemming which are used by information retrieval systems. 
In the contrary, linguistic approach uses more profound NLP techniques (part-of-speech, rhetoric relations, semantic, etc.) to generate summaries either by extraction or abstraction. 

\subsubsection{Topic words}

The presence of some words such as ``significant", ``impossible", etc. can be a strong indicator of the relevance of a sentence to the main topic \citep{69-edmundson}.
A dictionary can be prepared from a corpus to save three types of cue words: \textit{Bonus words} which are positively relevant, \textit{Stigma words} which are negatively relevant and \textit{Null words} which are irrelevant.
The score of a sentence  $ s_i $ based on this feature is the sum of the weight of every word $ w $ according to the dictionary, as indicated in Equation \ref{eq:edmundson-cue}.
\begin{equation}
\label{eq:edmundson-cue}
Score_{cue}(s_i) = \sum_{w \in s_i}{cue(w)}
\text{ where }
cue(w) = \left\lbrace 
\begin{array}{ll}
b > 0 & \text{if } (w \in Bonus) \\
\delta < 0 & \text{if } (w \in Stigma) \\
0 & otherwise 
\end{array} 
\right. 
\end{equation}

In \citep{09-abdelfattah-ren}, the cue words are divided into two groups: positive keywords and negative keywords.
Positive keywords are defined as ``the keywords frequently included in the summary". 
The score of a sentence $ s_i $ using positive keywords is given by Equation \ref{eq:abdelfattah-cue+}.
\begin{equation}
\label{eq:abdelfattah-cue+}
Score_{cue}(s_i) = \frac{1}{|s_i|} \sum_{w \in s_i}{tf(w) * P(s_i \in S | w)}
\end{equation}
Where: 
$ P(s_i \in S | w) $ is the probability that a sentence $ s_i $ belongs to a summary $ S $ given the occurrence of the word $ w $, which can be estimated using machine learning. 
Accordingly, negative keywords are the words unlikely to be in a summary, thus $ P(s_i \notin S | w) $.

\subsubsection{Indicators}

\begin{figure}[!ht]
	\begin{center}
		\includegraphics[width=.7\textwidth]{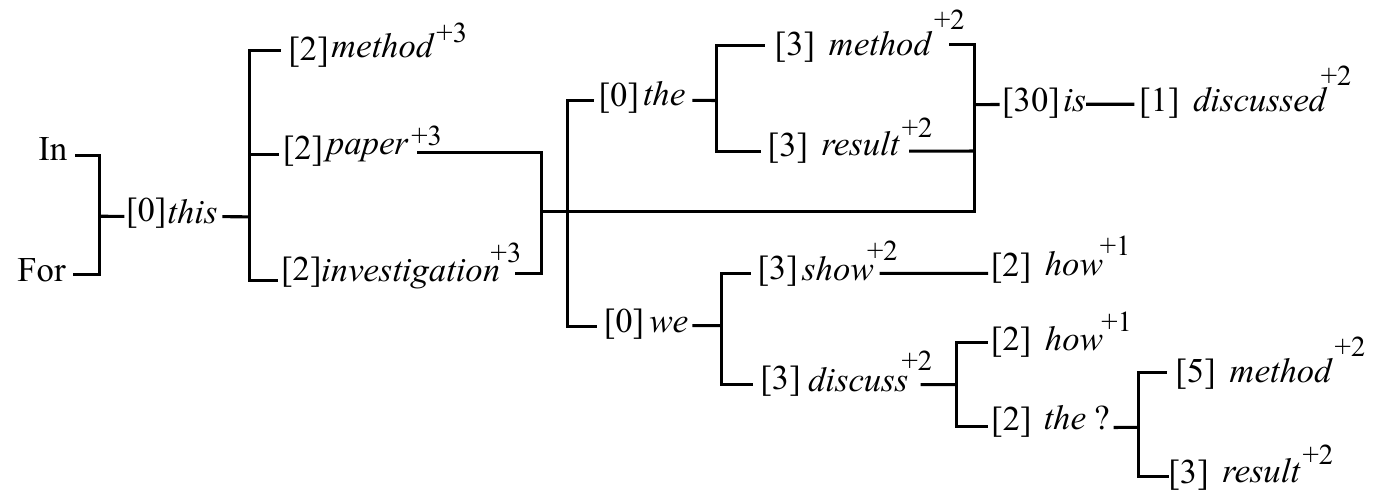} 
		\caption{A slightly simplified template \citep{81-paice}.}
		\label{fig:paice-template}
	\end{center}
\end{figure}

\citet{81-paice} defines them as ``\textit{commonly occurring structures which explicitly state that the sentences containing them have something important to say about the subject matter or the message of the document}";
For example, ``the principal aim of this paper is to investigate ...".
The identification of such indicators is not an easy task; in his work, the author followed these steps:
\begin{itemize}
	\item We can't list all the indicators due to their variations.
	For example, these expressions have the same structure: ``This article is concerned with ...", ``Our paper deals with ...", ``The present report concerns ..." and ``The following discussion is about ...". 
	So, the solution is to use some templates.
	
	\item Using templates, some sequences of words may show up which are not part of the indicators themselves.
	The solution is to use skip limits between the paradigms of a given template.
	
	\item There exists optional words, but can add a weight when be used, such as the word ``here" in the expression ``The purpose \textbf{here} is to ...". 
	This can be addressed by defining multiple paths in the template.
	
	\item To handle the variations of words, their stems can be used in the templates instead.
	
\end{itemize}
Figure \ref{fig:paice-template} represents a template, where the words or stems are paradigms.
The skip limits are shown thus: [3].
Weight increments are shown thus: +2.
A query (?) denotes an optional paradigm. 

\subsubsection{Co-reference information}

Some works use statistical approach to calculate the score, but they use linguistic techniques for that, so we can consider them as linguistic ATS systems.
In the work of \citet{07-orasan-stafford}, the authors try to use anaphora resolution to improve the  informativeness of summaries.
Sentences, usually, contain pronouns rather than words which lead to incorrect score calculation. 
Anaphora resolution will increase the frequencies of words referred by these pronouns, and produces more accurate frequency counts.
The authors use a simple term frequency algorithm to score sentences, and six anaphora resolution methods.
The average informativeness was improved using anaphora resolution.

Semantic representations of terms are often used to generate summaries. 
Using ontologies and lexicons like Wordnet \citep{1995-miller}, the semantic relationship between sentences' words can be exploited to enhance summary generation.
\citet{08-hennig-al} train an SVM classifier to identify salient sentences using ontology-based sentence features. 
This classifier is used to map the sentences to a taxonomy created by the authors, where each sentence is assigned a set of tags.
For each sentence, the tags are used to calculate the similarity with the document tags to determine how well a sentence represents the information content of its document in the ontology-space.

\subsubsection{Rhetorical structure}

\begin{figure}[!ht]
	\begin{center}
		\includegraphics[width=.5\textwidth]{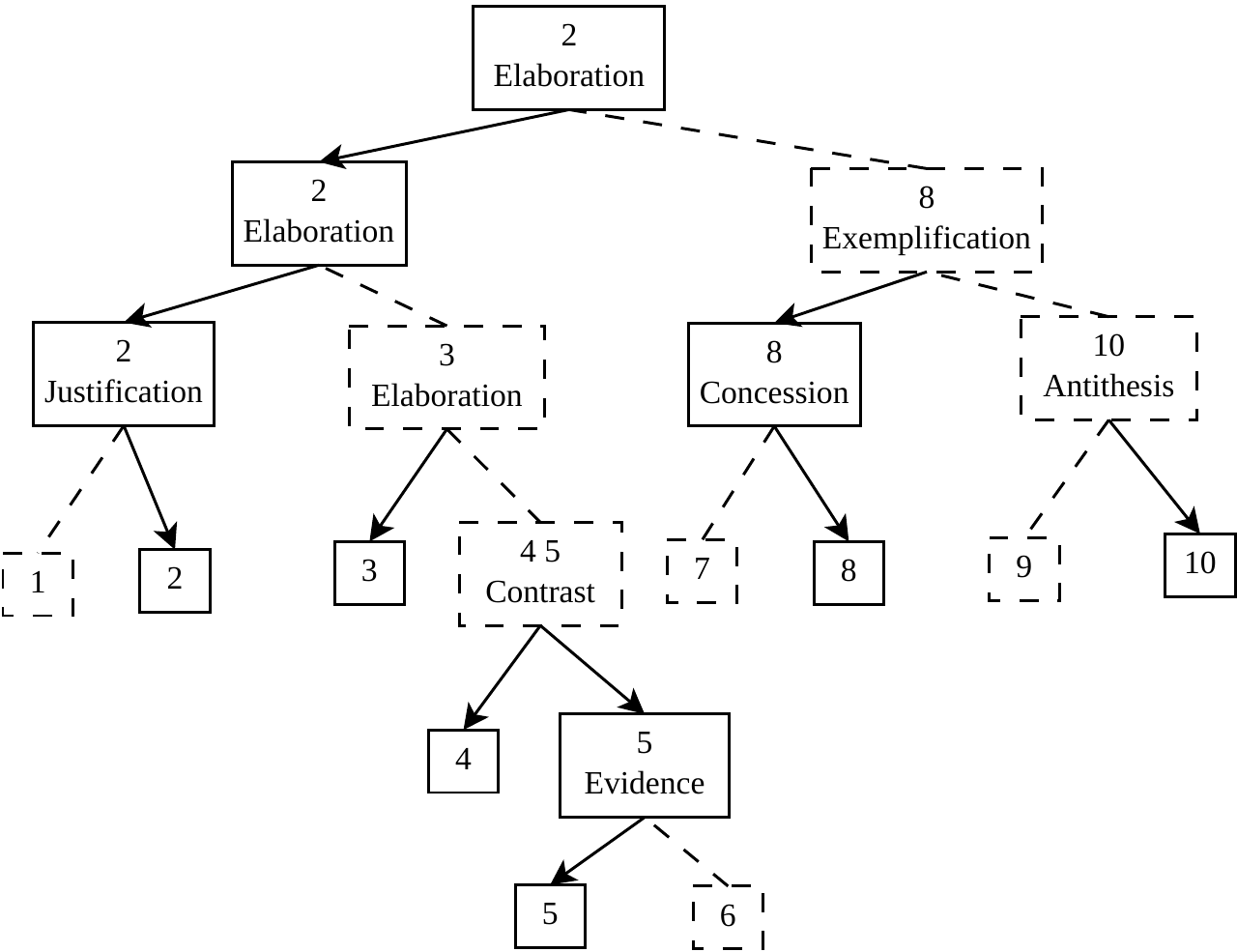} 
		\caption{An example of a rhetorical tree \citep{98-marcu}.}
		\label{fig:rhet-tree}
	\end{center}
\end{figure}

The structure of discourse can be exploited through rhetorical relations between sentences to generate summaries. 
\citet{94-ono-al} use a penalty score defined over different rhetorical relations to exclude non-important sentences.
In \citep{98-marcu}, a discourse tree is built to reflect the rhetorical relations between the text's sentences as illustrated in Figure \ref{fig:rhet-tree}, where the leafs are sentences.

The author uses seven metrics based on the rhetorical tree to find the best discourse interpretation similar to those of summaries: The clustering-based metric, the marker-based metric, the rhetorical-clustering-based metric, the shape-based metric, the title-based metric, the position-based metric and the connectedness-based metric.

\citet{14-kikuchi-al} propose a single document ATS based on nested tree structure. 
The authors exploit words dependency and rhetorical dependency  by constructing a nested tree composed of a document tree and a sentence tree. 
The document tree has sentences as nodes and head modifier relationships between sentences obtained by RST as edges. 
The sentence tree has words as nodes connected by head modifier relationships between them obtained by the dependency parser.
The summarization is formulated as combinatorial optimization problem in order to trim the nested tree.

A more recent work using RST is the work of \citet{16-goyal-eisenstein}, to fix the problems of local inference techniques which do not capture document-level structural regularities, and annotated training data.
So, the authors investigated the use SampleRank \citep{11-wick-al} structure learning algorithm as a potential solution to both problems.

\subsection{Machine learning (ML)}

Usually, machine learning approach is coupled with other approaches to improve and estimate their generation rules instead of fixing them manually.
It can solve the problem of combining features in statistical approach.
Many works have been conducted to generate summaries using machine learning, some focus on the choice of appropriate classification methods \citep{95-kupiec-al,02-osborne,05-yeh-al}, some try to solve the problems related to training phase, such as the absence of labeled corpora \citep{02-amini-gallinari}, others use heterogeneous features in their algorithms and try to combine them using machine learning \citep{08-wong-al,10-yatsko-al}, etc. 

\subsubsection{As a tuning function}

ML can be used with other approaches to tune certain parameters. 
It is mostly accompanied with statistical approach to solve the problem of fixing features weights, where these features scores are combined linearly into one score.
\citet{05-yeh-al} propose a method using genetic algorithm to estimate the features weights (sentence position, keywords, centroid and title similarity). 
The learning makes the system dependent to the corpus's genre. 
A solution is to tune the features according to the input document's genre as suggested in \citep{10-yatsko-al}. 
45 features are used to train the system on three genres: scientific, newspaper, and artistic texts. 
The system can detect the genre of the input document and execute the adequate model of scoring. 

\subsubsection{As a decision function}

Given a corpus of documents with their extractive summaries, a machine learning algorithm can be used to decide if a unit (sentence mostly) belongs to the summary or not.
\citet{95-kupiec-al} propose an ATS system based on Bayes classification, in order to calculate the probability that a summary may contain a given sentence.
So, for each sentence $ s_i $, the probability that it belongs to a summary $ S $ using a features vector $ \overrightarrow{f} $ is given in Equation \ref{eq:bayes-kupiec}.
\begin{equation}
\label{eq:bayes-kupiec}
P(s_i \in S | \overrightarrow{f}) = %
\frac{\prod_{j = 1}^{|\overrightarrow{f}|} P(f_j | s_i \in S) * P(s_i \in S)}
{\prod_{j = 1}^{|\overrightarrow{f}|} P(f_j)}
\end{equation}
Where: 
$ P(s_i \in S) $ is a constant (so it can be emitted, since the probability is used for reordering), $ P(f_j | s_i \in S) $ and $ P(f_j) $ can be estimated from a corpus.
Choosing the right classification algorithm is another issue; for instance Bayes classification supposes features independence, which is not always the case.
From this point of view, \citet{02-osborne} uses a maximum entropy based classifier which, unfortunately, performs lower than Bayes' because it tends to reject a lot of sentences; but when he adds a prior probability, the method performs well and surpasses Bayes-based method.
His method, though, is not used to reorder the sentences using the probability, but to classify them into two classes: summary sentences and other sentences.

\subsubsection{Bayesian topic models}

Topic models are based on identifying the main concepts from the input document(s) and find the relationship between them to construct a hierarchy. 
Words of each input document are assigned to a given number of topics, where a document is a mixture of different topics and each topic is a distribution of words. 
Bayesian topic models are quite sophisticated for multi-document summarization since they make difference between documents in contrast to most ATS methods which consider them as one giant document \citep{11-nenkova-mckeown}.
One of Bayesian topic models' advantages is their simplicity to  to incorporate different features, such as cue phrases used in \citep{08-eisenstein-barzilay} for unsupervised topic segmentation.

\begin{figure}[ht]
	\begin{center}
		\includegraphics{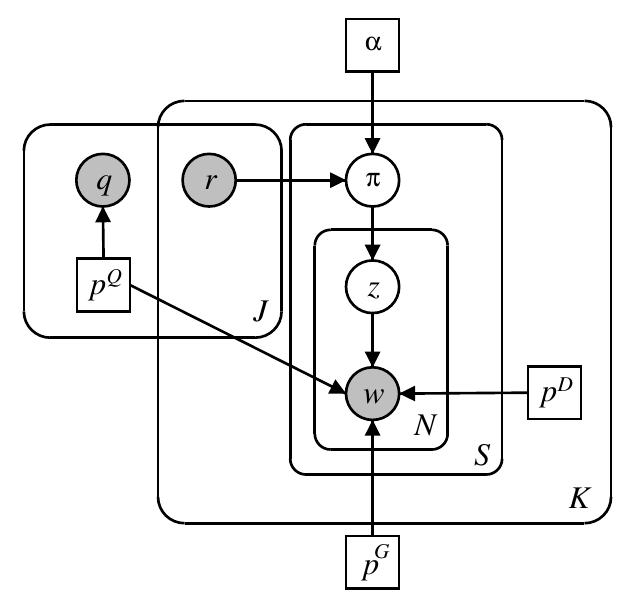} 
		\caption{Bayesian Query-Focused Summarization Model proposed in \citep{06-daumeiii-marcu}.}
		\label{fig:btm-daumeiii}
	\end{center}
\end{figure}

\citet{06-daumeiii-marcu} propose a query-driven muli-document summarization method using Bayesian topic models which is reported in \citep{11-nenkova-mckeown} as one of the first works in this direction.
In their method, the authors train their system to have three probability distributions: general English learning model ($ P^G $), background document language model ($ P^D $) for each one of the $ K $ available documents with $ S $ number of sentences and $ N $ number of words $ w $ in a sentence, and query language model ($ P^Q $) for each query $ q $ over a set of $ J $ queries.
Using this trained model, they estimate $ \pi $ using Expectation propagation \citep{01-minka}.
Figure \ref{fig:btm-daumeiii} represents the graphical model of their method, where $ r $ is the relevance judgment,  $ z $ is the word-level indicator variables, $ \pi $ is the sentence level degrees, and $ \alpha $ is a Dirichlet distribution as prior over $ \pi $.

\citet{11-celikyilmaz-hakkani} introduces two-tiered topic model (TTM) which models topics as two levels: low-level topics which are distributions over words, and top-level topics which are correlations between these lower level topics given sentences.
One problem with TTM is its incapability to differentiate general words from specific ones given a query. 
A Sentence containing general words, which are more frequent in document clusters, must have more chance to be included into the summary. 
The authors present an enriched TTM (ETTM) generative process to sample words from high-level topics leading to three words distributions: low-level topics over corpus specific words, high-level topics over corpus general words, and background word distributions.

\citet{15-yang-al} argue that using expert summaries to indicate only two levels of topics limits the practical applications of \citet{11-celikyilmaz-hakkani}'s model, because we can draw many latent topics from multiple documents.
The availability of modal summaries and their quality as golden standards are two other challenges to this method. 
Furthermore, TTM does not take word dependency in consideration. 
To address these limits, they propose a new method called contextual topic model (CTM) which is based on Bayesian topic model (TTM), hLDA model \citep{10-blei-al}, topical N-grams \citep{07-wang-al}, with some concepts of graph models. 

\subsubsection{Reinforcement learning (RL)}

\citet{12-ryang-abekawa} see ATS extractive approach as a search problem. 
They model the problem of constructing a summary as a RL and attempt to optimize the summary score function based on some feature representation, where the  score is a trade-off between relevance and redundancy.
The summary is represented as a state with a history of actions and an indicator showing if it is terminal or not.
An action (inserting a sentence into the summary or finishing) is selected based on a probability calculated using Boltzman selection.
As for the reward, the summary is rewarded only when it is in finish state; It is rewarded by the summary score if it does not violate the size limit, and it is penalized otherwise.

\citet{15-henss-al} follow the same method but with some changes. 
They use a different reward function based on reference summaries during training. 
Instead of using temporal difference learning, they use Q-Learning to determine the value of the partial summary and the value of adding a new sentence to a summary state.
Their method learns one global policy for a specific summarization task instead of one policy for each document cluster.

Similarly, \citet{15-rioux-al} base their work on the method of \citet{12-ryang-abekawa}. 
They use an improved version of TD called SARSA which, in addition to modeling state space, models actions space too. 
Their reward function is immediate at every action to help the learner get immediate feedback.

\subsubsection{Deep learning}

Deep learning has gained a lot of attention recently, even in ATS research.
It is based on large neural networks (NN) which, eventually, needs a huge amount of data to be trained. 
\citet{15-rush-al} are the first to successfully apply deep learning to an abstractive ATS. 
The method uses a local attention-based model to generate each word of the summary conditioned on the input sentence.
Their system called NAMAS\footnote{NAMAS code: \url{https://github.com/facebookarchive/NAMAS} [03 October 2017]} performs well on ``DUC-2004 shared task" corpus despite being tested using ROUGE metric which, mostly, encourages extractive summaries. 
Their method shows some limitations when it comes to input document and summary's sizes. 
It processes only documents with a size of about 500 words and produces a very short summary (about 75 characters). 

Following their lead, \citet{16-nallapati-al} also use an Attention model in the encoder-decoder.
When they decode, they use only the words that appear in the source document following a method called ``Large Vocabulary Trick" (LVT) \citep{15-jean-al}. 
Then to introduce new words, they add a layer of ``word2vec nearest neighbor" in the input. 
The decision whether to use a word from the input or a new word based on the context is guaranteed by another layer they call ``Switching Generator/Pointer" layer \citep{15-luong-al,15-vinyals-al}.

\citet{17-ling-rush} try to fix the problem of speed when long source sequences (document summarization) are processed using sequence-to-sequence models with attention.
So, they propose to use a two layer hierarchical attention; Where the first layer's function is to select one or more important chunks from the input document using hard attention then feed it/them into the second layer using sequence-to-sequence model.
They use reinforcement learning to train the hard attention model.
the method shows promise to scale up existing methods into large inputs, but fails to beat the standard sequence-to-sequence model. 

Deep learning  was not used just for abstractive ATS, but also for extractive one.
In \citep{12-liu-al,15-zhong-al}, the authors propose a query-oriented multi-document ATS method based on a deep learning model with three layers: concepts extraction, summary generation, and reconstruction validation. 
Concentrated information is used with dynamic programming to seek most informative sentences. 

\citet{14-denil-al} use a hierarchical ConvNet architecture (CNN) divided into a sentence level and a document level.
The sentence level learns to represent sentences using their words and the document level learns to represent the input document using the first level. 
Then these representations are used to score how pertinent a sentence can be towards the entire document.

In \citep{15-cao-al}, a Recursive Neural Networks (R2N2) is used to rank sentences for multi-document summarization. 
They use two types of hand-crafted features as inputs: word features and sentence features. 
This enables the system to learn how to score sentences based on a hierarchical regression process which uses a parsing tree to score sentence constituents such as phrases. 

Like \citet{15-zhong-al}, \citet{17-yousefiAzar-hamey} propose an extractive query-oriented summarization method based on deep learning but for single-document ATS.
The main difference is the use of an unsupervised approach with deep auto-encoder (AE) which was used in \citet{15-zhong-al} as a word filter rather than a sentence ranker. 
The AE, in this case, can learn different features rather than manually engineering them.
Furthermore, the authors investigate the effect of adding random noise to the local word representation vector on the summary.

In \citep{17-ren-al}, a neural network of two levels using contextual relations among sentences is proposed to improve sentence regression's performance. 
The first level captures sentence representations using a word-level attentive pooling convolutional neural network. 
Then, the second one constructs context representations using sentence-level attentive pooling recurrent neural network. 
These learning representations and the similarity of a sentence with its context are used to extract useful contextual features. 
The authors train the system to learn how to score a sentence to fit the ground truth of ROUGE-2.

\subsection{Compression}

The compression or reduction of sentences is the elimination of non informative parts, especially when the sentence is too long.
\citet{99-jing-mckeown} confirm that the compression is often used by professional summarizers. 
In fact, they found out that 78\% of summaries sentences are borrowed from the original document, and that half of them have been compressed.

In \citep{02-knight-marcu}, two statistical compression methods have been proposed: using the noisy channel model, and using decision trees.
The first method supposes that the input sentence was short but has been bloated with additional words. 
For a given long sentence, a syntactic tree is produced to which a quality score is attributed.
It is calculated using probabilistic context free grammar (PCFG) scores and the probabilities of next sentence calculated with bi-gram language model. 
They measure the likelihood of transforming a wide tree to a reduced one using statistics gathered from a corpus of documents and their summaries.
The second method is based on decision trees used to learn the different rules to reduce sentences.

Likewise, in their statistical compression method, \citet{07-Zajic-al} see the sentence as if it was a headline distorted by adding other words to it. 
They use Hidden Markov Models (HMM) to represent the problem and finding the headline which maximizes the likelihood of this headline generating the sentence. 
This likelihood is estimated using probabilities that a word is followed by another in a corpus of English headlines taken from TIPSTER corpus. 
The authors propose another rule-based method where they use a parser to get the syntactic representation of the sentence, used to remove some components such as temporal expressions, modal verbs, complimentizer \textit{that}, etc.
The compression is executed iteratively by removing one component at a time, outputting the compressed sentence into the extraction module.

Other methods have been proposed to improve sentences compression using not only one sentence but using similar sentences.
In \citep{08-cohn-lapata}, instead of shortening the sentence by removing words or components, the authors introduce additional steps as substitution, reordering, and insertion.
In \citep{10-filippova}, a method is proposed to summarize a group of related sentences to a reduced one.
In this method, the sentences are represented by one graph which is used to generate a reduced summary by following the shortest path.

Deep learning has been applied in order to compress sentences. 
\citet{15-filippova-al} use Long Short Term Memory models (LSTM) in order to perform a deletion-based sentence compression. 
The method performs very well either in term of readability or informativeness even without affording syntactic information (PoS, NE tags and dependencies).
The same method is used in \citep{17-hasegawa-al} to compress Japanese sentences, but with some changes. 
These modifications are based on three Japanese language characteristics: frequent verbs are nominalized and nouns are abbreviated, non verbs can be the root node, and easily estimated subjects and objects are omitted.

\subsection{Approaches discussion}

We based our taxonomy on resources dependency: is a method based on some corpora? does it depend on some language toolkits? how much calculation power it needs?
Our intension is to classify ATS methods based on resource availability and the effort spent to be implemented. 
Also, we try to show the capacity of each approach to support multilingualism.

Statistical features based methods were the first introduced in ATS. 
Features like term frequency, position and length are language independent and are good indicators of sentence relevance. 
A statistical approach does not need much language dependent tools; just some basic NLP tasks such as sentence segmentation, tokenizaion, stop word elimination, and stemming. 
Mostly, it is easy to be implemented and does not need a lot of processing power.
But, mostly these features are combined together in hope to increase sentence relevance. 
This leads to a more complicated problem which is how to combine them, and how much amount of influence of each one. 
The problem can be solved by combining them linearly, and fix their weights through experiments using a corpus or estimate them using machine learning.
Either way, we will have another problem which is language dependency. 

Graph-based approach seeks to exploit the shared information between sentences. 
It is a bottom-up method which discusses the similarity problem from the perspective of content structure \citep{15-yang-al}. 
It can be language-independent when using just lexical similarities (mostly, cosine similarity) and statistical features. 
But, some room for improvement can be done by considering more language dependent similarities like semantic similarity. 
Also, the modal can be improved using machine learning, by introducing prior probabilities into the equation, as in \citep{08-liu-al}.
Processing a great amount of text using iterative graphs can consume more processing prower; this should be investigated to better understand how far this approach can go.

Linguistic approach is more powerful than the statistic one, because it integrates richer processing of the input text. 
Also, it can be used either for extractive or abstractive summarization.  
The problem with this approach is the lack of appropriate NLP tools for certain languages. 
It can be as simple as using sentence components (verbs, nouns, etc.) as statistical features, or as hard as using sentence structure and its relationships with others to generate a new text. 
Mostly, it is harder to be implemented and takes more time to generate a summary. 
\citet{11-nenkova-mckeown} suggests using linguistic methods as a post-processing task to improve linguistic quality of the generated summary rather than a processing one.
According to authors, it is unclear how much this approach can improve content selection compared to the methods using no linguistic relations. 

All approaches discussed previously can be ameliorated using machine learning (ML).
But in the other hand, they can loose multilingual property unless the system is trained on as many languages as possible. 
Still the problem of corpus domain; training your system on news articles does not mean it can handle fiction as good as it does with news.
Also, collecting labeled corpora for one language can be a very hard work, let alone many languages. 
Some works seeks to use unlabeled data, and some propose auto-supervised methods such as in \citet{02-amini-gallinari}.

Sentence compression seeks to get rid of non informative parts allowing us to have shorter sentences. 
It can be used as a post-processing task after generating a summary, which will eliminate more redundancy and allows more space for other sentences to be included. 
It is a language dependent task which can't be used alone to have a summary, but in conjunction of other approaches.

We tried to summarize all the above discussion in Table \ref{app-comp}. 
The table shows some advantages and disadvantages of each approach, some encountered problems and the eventual fixes.
\begin{table}[ht]
	\caption{Comparison between different ATS approaches.\label{app-comp}}
\begin{tabular}{p{.12\textwidth}p{.25\textwidth}p{.25\textwidth}p{.25\textwidth}}
	\hline\hline\noalign{\smallskip}
	
	Approach & pros (+) \& cons (-) & problems & fixes \\
	\noalign{\smallskip}\hline\noalign{\smallskip}
	
	Statistical 
	& 
	+ less resources \newline 
	+ simple \& fast \newline
	- readability
	& 
	* features combination \newline
	* relevancy \newline 
	* redundancy in sentences
	& 
	* machine learning \newline
	* linguistic features \newline 
	* compression
	\\
	
	\hline\noalign{\smallskip}
	
	Graph 
	& 
	+ less resources \newline 
	+ simple \newline
	+ coherence \newline
	- processing power
	& 
	* sentences similarity \newline 
	* sentences \& documents properties
	& 
	* tf-idf, semantic \newline 
	* statistical \& linguistic features, sentence-document similarity, temporal property
	\\
	
	\hline\noalign{\smallskip}
	
	Linguistic 
	& 
	+ more accurate  \newline 
	+ abstractive ATS  \newline 
	- resources (toolkits) \newline 
	- complex \newline
	- processing power
	& 
	* generation rules
	&
	* machine learning 
	\\
	
	\hline\noalign{\smallskip}
	
	Machine learning 
	& 
	+ deducing rules automatically \newline
	- lack of corpora \newline
	& 
	* labeled corpora \newline
	* corpora insufficiency
	& 
	* reinforcement learning \newline
	* corpora creation
	\\
	
	\hline\noalign{\smallskip}
	
	Compression 
	& 
	+ less redundancy \newline
	- sentence level
	& 
	* sentence level
	& 
	* fuse with other approaches as a post-processing task
	\\
	
	\noalign{\smallskip}\hline\hline
\end{tabular} 
\end{table}

\section{Evaluation}
\label{sec:eval}

Evaluating summarization systems is one of the most difficult tasks. 
It is not clear which is the perfect summary, since humans produce different summaries and yet we can consider them all good. 
Also, the evaluation must be accurate, objective and fast. 
This led to the appearance of many evaluation methods which can be automatic, manual or semi-automatic. 
Many workshops have been organized to evaluate the performance of summarization methods, and test how effective evaluation methods are.

\subsection{Evaluation methods}

\begin{figure}[!ht]
\begin{center}
	\includegraphics[width=.75\textwidth]{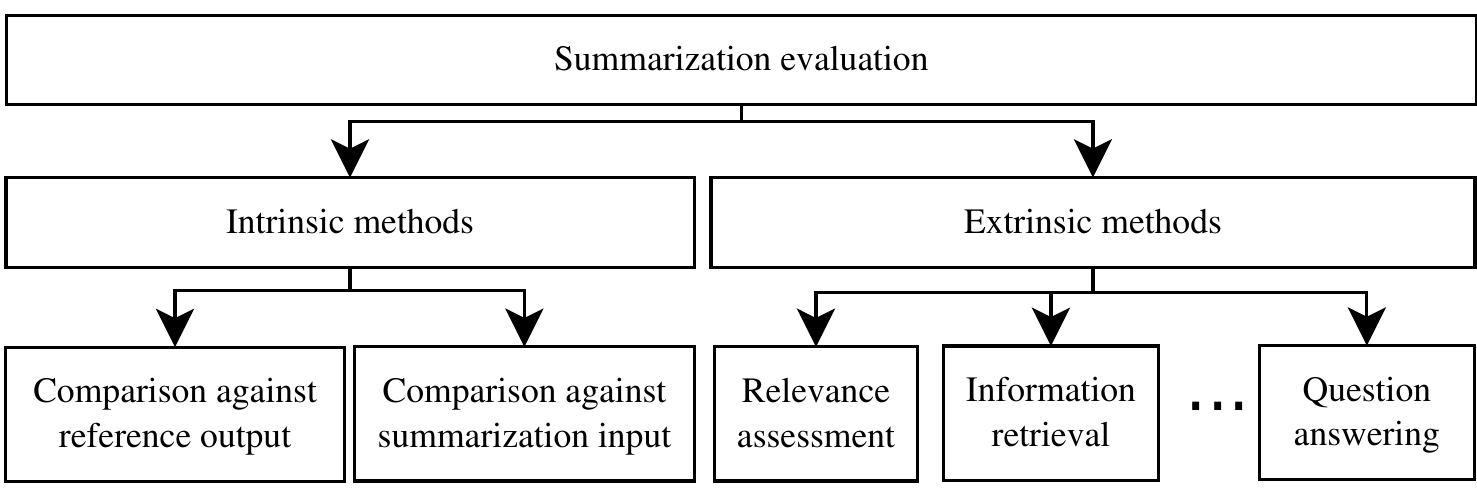} 
	\caption{Classification of evaluation methods.}
	\label{fig:eval-classif}
\end{center}
\end{figure}

ATS evaluation methods can be considered as intrinsic or extrinsic \citep{01-mani}.
Intrinsic evaluation is meant to evaluate the system in of itself. 
It is based on two criteria: summary coherence and summary informativeness. 
In the other hand, extrinsic evaluation seeks to determine the effect of summarization on other tasks. 
Figure \ref{fig:eval-classif} represents the different categories of summarization evaluation, as described in \citep{01-mani}.

\subsubsection{Intrinsic evaluation}

The evaluation can be done by comparing the generated summary to the original text or to a reference summary. 
When the system's summary is compared to the original text, actually, we look for the quantity of information recovered; more like information retrieval systems.
Comparing it to a reference summary will allow us to quantify how good the system can be against humans.
Also, the intrinsic methods can be seen as: text quality evaluation methods or content evaluation methods \citep{12-steinberger-jezek}.
Text quality evaluation methods attempt to verify linguistic aspects of the generated summary such as grammaticality, reference clarity and coherence. 
The content evaluation methods can be divided into two sub-classes: co-selection methods which evaluate the system based on selecting the right sentences, and content based methods which go deeper by using smaller units such words and n-grams.
 
Given a summarizer which takes an input document $E$ and generates a summary $S$, and given a reference (model) summary $M$ which is some preselected sentences from $ E $; we can define the following relations: 
TP (true positive), TN (true negative), FP (false positive) and FN (false negative).
The recall is the quantity of right information recovered by a system comparing to what the it should recover (see Equation \ref{eq:recall}).
\begin{equation}
\label{eq:recall}
R = \frac {TP} {TP + FN} = \frac {M \cap S} {S}
\end{equation}
The precision is the quantity of right information recovered by a system comparing to what it has recovered (see Equation \ref{eq:precision}).
\begin{equation}
\label{eq:precision}
P = \frac {TP} {TP + FP} = \frac {M \cap S} {M}
\end{equation}
The F-Score is a mix between the recall and the precision using harmonic mean (see Equation \ref{eq:f-score}). 
If $ \beta $ is more than one (1), the recall is advantaged. 
If it is less than one, the precision is advantaged. 
The most used F-Score is F1-score which is a trade-off between recall and precision.
\begin{equation}
\label{eq:f-score}
F_{\beta} = \frac
{(1+\beta^2) P \times R}
{ \beta^2 P + R}
\end{equation}
We have to say, this type of evaluation does not give systems credit if they choose other sentences similar to the preselected ones. 
It may be useful to use it when the summary must contain a precise information from the input text. 
But, in general, to evaluate a summarization system it must be avoided, especially for abstractive methods.

``\textit{Recall-Oriented Understudy for Gisting Evaluation}" (ROUGE)\footnote{ROUGE: \url{http://www.berouge.com/Pages/default.aspx} [23 November 2016]}, proposed by \citet{03-Lin-hovy}, is a method inspired from another used for automatic translation evaluation called ``\textit{BiLingual Evaluation Understudy}" (BLEU) \citep{02-papineni-al}. 
The objective is to automatically measure the quality of the generated summary comparing to a reference one.
The idea is to compute the number of units (N-grams) in both the system's summary and the reference one and calculate the recall.
Since a text can have many summaries, this method allows the use of many reference summaries.
Many variances of ROUGE have been proposed \citep{04-lin}: ROUGE-N, ROUGE-L, ROUGE-W, ROUGE-S, et ROUGE-SU. 
Equation \ref{eq:rouge-n} describes how ROUGE-N is calculated.
\begin{equation}
\label{eq:rouge-n}
ROUGE-(N) = \frac{\sum_{S \in Summ_{ref}}{\sum_{N-gram \in S}{Count_{match} (N-gram)}}}
{\sum_{S \in Summ_{ref}}{\sum_{N-gram \in S}{Count (N-gram)}}}
\end{equation}
Where, \textit{N} is the size of the N-gramme, 
$count_{match}(N-gram)$ is the number of N-grammes found in both the candidate and the reference summaries. 
$Count (N-gram)$ is the number of N-grammes in the reference summary. 

Using similarity measure between automatic summary and human-made one can present some limitations \citep{07-nenkova-al}:
human variation, analyze granularity, semantic equivalence, semantic equivalence and the comparison between extractive and abstractive summaries.
%
%
%
%
Pyramid \citep{07-nenkova-al} comes to solve these limitations using a semi-automatic approach. 
Given a number of reference summaries, annotators are asked to define summary content units (SCUs) where the weight of each SCU can be calculated as the number of relative reference summaries. 
The process of defining the SCUs is not given, they can be as small as modifiers of a noun phrase or as large as a clause. 
The SCUs are regrouped in a pyramid of $ n $ tiers, where $ n $ is the maximum weight of SCUs. 
So an SCU which belongs to the tier $ T_i $ will have a score of $ i $, where $ |T_i| $ is the number of SCUs in this tier.
For a summary with $ X $ SCUs, Equation \ref{eq:pyramid} shows the optimal content score, where SCUs which do not belong to the pyramid will have a zero weight.
\begin{equation}
\label{eq:pyramid}
\max = \sum\limits_{i=j+1}^{n} i * |T_i| + j * (X - \sum\limits_{i=j+1}^{n} |T_i|),
\text{ where } j = \max\limits_i (\sum\limits_{t = i}^{n} |T_t| \ge X)
\end{equation}

Basic Elements (BEs) are minimal semantic units which can be extracted from a sentence \citep{06-hovy-al}. 
Their goal is to automate summaries evaluation in contrast of Pyramids which needs human involvement. 
This later can introduce some problems such as human variability and evaluation expensiveness in time and cost.
To evaluate a summary, three different modules are proposed: BE breakers, BE matchers and BE scorers. 
The first one is used to break the text into BEs which are defined by \citep{06-hovy-al} as:
``\textit{1- The head of a major syntactic constituent (noun, verb, adjective or adverbial phrases), expressed as a 	single item, or, 2- A relation between a head-BE and a single dependent, expressed as a triple (head | modifier | relation)}".
The matching of BEs is either lexical, using lemmas, using synonyms from Wordnet \citep{95-miller}, using paraphrases or using semantic generalization.
Each BE gets a weight of 1 point for each related reference summary.

AUTOmatic SUMMary Evaluation based on N-gram Graphs (AutoSummENG) \citep{06-giannakopoulos-al} attempts to be language neutral, fully automatic and context sensitive.
The method is based on n-grams, so to be language neutral no preprocessing must be done, this is why the n-grams are extracted over the characters.
Then, a graph is constructed where the n-grams represent its nodes and each arc connecting two n-grams is the number of times these n-grams are judged as neighbors given a distance window $ D_w $. 
Given two summaries $ S_i $ and $ S_j $ with a set of graphs $ \mathbb{G}_1 $ and $ \mathbb{G}_2 $ respectively, and given two graphs $ G^i \in \mathbb{G}_1 $ and $ G^j \in \mathbb{G}_2 $ with the same rank of n-grams, the value similarity is given in Equation \ref{eq:val-sim}.
\begin{equation}
\label{eq:val-sim}
VS(G^i, G^j) = 
\frac{\sum_{e \in G^i} \left(\mu(e, G^j) * \frac{min(w_e^i, w_e^j)}{min(w_e^i, w_e^j)}\right)}{max(|G^i|, |G^j|)}
\end{equation}
With $ \mu $ as the membership function, which returns 1 when $ e \in G^j $ and 0 otherwise.
$ w_e^i,\ w_e^j $ are the weights of the edges of the element $ e $ in the graphs $ G^i,\ G^j $ respectively.
The overall similarity of $ \mathbb{G}_1 $ and $ \mathbb{G}_2 $ is the weighted sum of the $ VS $ over all ranks (Equation \ref{eq:oval-sim}).
\begin{equation}
\label{eq:oval-sim}
VS^O(\mathbb{G}_1, \mathbb{G}_2) = 
\frac{\sum_{r \in [L_{min}, L_{max}] } r * VS^r }{\sum_{r \in [L_{min}, L_{max}]} r}
\end{equation}
Where $ r $ is the rank of n-grams; 
$ L_{min},\ L_{max} $ are the minimum and the maximum rank of n-grams; 
$ VS^r $ is the value similarity using the graphs with n-gram of rank $ r $.
The grade of a summary is the average of all its similarities with the reference summaries. 
Merged Model Graph (MeMoG) is somehow a variant of AutoSummENG, where all the graphs of reference summaries are merged into one graph \citep{10-giannakopoulos-karkaletsis}.

N-gram graph Powered Evaluation via Regression (NPowER) \citep{13-giannakopoulos-karkaletsis} is a combination of many of the past metrics via optimization. 
The idea is to create a method which stands as an independent judge to combine the different metrics grades into a better estimate. 
Machine learning (linear regression) is used to estimate the final grade. 
Given a vector of features $ \overline{x} \in \overline{X}  $ (here, ROUGE, BE, etc.) and a target numeric feature $ y \in \mathbb{R},\ y = f(\overline{x}) $, where $ f $ is an unknown function (human scores or Pyramid scores), the idea is to estimate a combination function $ \tilde{f} $ as shown in Equation \ref{eq:npower}.
\begin{equation}
\label{eq:npower}
\tilde{f}: \overline{X} \rightarrow \mathbb{R}: 
\sum (\tilde{f}(\overline{x}) - f(\overline{x}))^2 \rightarrow 0,\ \forall \overline{x} \in \overline{X}
\end{equation}

Looking to the expansiveness of the summarization systems evaluated by these past metrics, we can say that they are all background systems. 
But, how about just-the-news systems? In case we want to evaluate automatic summaries in term of new information which they contain.
Nouveau-ROUGE \citep{11-conroy-al} is a method based on ROUGE metrics to evaluate update summaries. 
The idea is to calculate ROUGE score between an original summary and an update one, then a high ROUGE score indicates high redundancy.
In TREC temporal summarization task \citep{13-aslam-al}, many metrics are proposed to address this issue. 
Given an event (Wikipedia event pages), gold standard updates called nuggets are extracted  and annotated to form a set of nuggets $ \mathcal{N} $ such as a nugget $ n \in \mathcal{N} $ has the properties: the time-stamp of revision history $ n.t $ and the importance provided by assessors (0: no importance to 3: high importance) $ n.i $. 
The evaluation metrics are designed to measure the degree to which a system can generate these nuggets in a timely manner.
Each system must generate a summary $ \mathcal{S} $ containing some updates, where an update $ u $ is a sentence length text ($ u.string $) having a time-stamp $ u.t $.
The earliest update $ u $ that match a nugget $ n $ is defined as $ M(n, \mathcal{S}) = \text{\textit{argmin}}_{u \in \mathcal{S}: u \approx n} u.t $.
Then, the inverse function is defined as $ M^{-1}(u, \mathcal{S}) = \{ n \in \mathcal{N}; M(n, \mathcal{S}) = u \} $.
The first metric is \textit{Expected Gain metric} (EG) which is similar to traditional notions of precision in IR evaluation (see Equation \ref{eq:eg}). 
When using latency discount, the chosen latency step was $ \alpha = 3600 * 6$ (6 hours).
\begin{equation}
\label{eq:eg}
EG(\mathcal{S}) = \frac{1}{|\mathcal{S}|} \sum\limits_{u \in \mathcal{S}} G(u, \mathcal{S})
\end{equation}
Where, 
\[
G(u, n) = \sum\limits_{n \in M^{-1} (u, \mathcal{S})} R(n) * \text{\textit{discounting factor}}
\]
\[
R(n) = \left\lbrace\begin{tabular}{ll}
Graded:& $ \frac{e^{n.i}}{e^{\max_{n' \in \mathcal{N}} n'.i}} $ \\
Binary:& $ 1 $ if  ($ n.i > 0 $), $ 0 $ otherwise
\end{tabular}\right.
\]
\[
\textit{discounting factor} = \left\lbrace\begin{tabular}{l}
Discount-free gain: $ 1 $ \\
Latency-discounted gain: $ L(n.t, u.t) = 1 - \frac{2}{\pi} \arctan(\frac{u.t - n.t}{\alpha}) $ 
\end{tabular}\right. 
\]
Another metric is \textit{Comprehensiveness metric} C(S) which is similar to traditional notions of recall in IR evaluation (see Equation \ref{eq:cs}).
\begin{equation}
\label{eq:cs}
C(S) = \frac{1}{\sum_{n \in \mathcal{N}} R(n)} \sum\limits_{u \in S} G(u, S)
\end{equation}

\subsubsection{Extrinsic evaluation}

ATS is, often, used to complete other tasks like executing instructions, information retrieval, question answering, relevance assessment, etc. 
To evaluate how well a summarization system is doing leads to evaluate its effectiveness towards its related task. 
In the relevance assessment task, the assessors try to score how well a summary is related to a given subject.
Another example is reading comprehension task where a human judge is asked to answer some questions based on the original document, then on its summary.
The correct answers number is considered as the score of this summary.

In TIPSTER SUMMAC evaluation \citep{99-mani-al}, two tasks are proposed to evaluate the impact of ATS on real world tasks: ad-hoc task and categorization task. 
In \textit{ad-hoc task}, the focus is to test the pertinence of indicative summaries towards a particular topic.
Given a document (summary or source text), a human subject is asked to determine its relevance to a given topic description, ignoring whether it was a full text or a summary.
The accuracy of the subjects is measured on how well they can indicate the relevance between the subjects and their relative full texts. 
Then the recall, precision and F1-score are calculated for the participants systems. 
In \textit{categorization task}, the purpose is to determine if a generic summary contains enough information to allow an analyst categorizing a document as quickly and correctly as possible. 
The evaluation is proceeded as in ad-hoc task's.
But, the human subject, after reading the document, has to choose one category out of five or choose ``None of the above". 
Then the three measures: recall, precision and F1-score are calculated.


To assess the usefulness of a summary, according to \citet{05-dorr-al}, the decision made by a human judge (subject) based on the summary must be compared to its own decision made on the full-text rather than to a gold standard.
This is motivated by the fact that the users judgments based on the original texts are more reliable than basing on gold standard judgments.
Given a summary/document pair $ (s,\ d) $, the function $ j(s,\ d) $ equals to:
\begin{itemize}
	\item $ 1 $: if the subjects have the same judgment on both $ s $ and $ d $.
	\item $ 0 $: if the subjects change their judgment between $ s $ and $ d $.
\end{itemize}
Equation \ref{eq:rel-pred} calculates the Relevance-Prediction score for a set of summary/document pairs $ DS_i $ in association with an event $ i $.
\begin{equation}
\label{eq:rel-pred}
\text{\textit{Relevance-Prediction}} (i) = \frac{\sum_{(s, d) \in DS_i} j(s,\ d)}{|DS_i|}
\end{equation}

Another relevance prediction example is TREC Real-Time Summarization (RTS\footnote{RTS evaluation: \url{http://trecrts.github.io/TREC2016-RTS-guidelines.html} [20 December 2016]}) track.
The purpose of summaries is to afford the users with tweets that are relevant to their profiles, and are novel.
Given a profile and a set of retrieved tweets, the gain $ G(t) $ of each tweet $ t $ is 0 if the tweet is not relevant, 0.5 if it is relevant or 1.0 if it is highly relevant.
This gain is attributed by some users having the target profile.
Based on this, three metrics are defined: Expected gain (EG), Normalized Cumulative Gain (nCG) and Gain Minus Pain (GMP) which are described in Equation \ref{eq:rts}.
\begin{equation}
\label{eq:rts}
\begin{tabular}{lllll}
$ EG = \frac{1}{N} \sum G(t) $ &,&
$ nCG = \frac{1}{Z} \sum G(t) $ &,&
$ GMP = \alpha * nCG - (1- \alpha) * P$
\end{tabular}
\end{equation}
Where $ Z $ is the maximum possible gain (given the ten tweet per day limit);
$ N $ is the number of tweets returned; 
$ P $ (pain) is the number of non-relevant tweets that are pushed, and $ \alpha $ controls the balance between the gain and the pain (0.33, 0.5, and 0.66 are used).

\subsection{Workshops and evaluation campaigns}

\subsubsection{TIPSTER SUMMAC}

It was launched in may 1998 by the US government, in order to evaluate ATS systems in a large scale.
Three evaluation tasks were defined, two extrinsic (adhoc and categorization tasks) and one intrinsic (question-answering) \citep{99-mani-al}:
\begin{itemize}

\item \textit{The ad-hoc task:} It is intended for indicative summaries based on a specific topic. 
Given a document (the evaluator do not know if it is a summary or a full text) and a description of a topic, the evaluator is asked to determine if this document is pertinent to the topic. 

\item \textit{The categorization task:} Its aim is to measure the effectiveness of a generic summary (ignoring the topic) to afford enough information allowing an analyst to categorize a document as quickly and correctly as possible. 
Given a document (the evaluator do not know if it is a summary or a full text), the evaluator must choose from five categories the one which is pertinent to the document, otherwise he choose "\textit{No category}".

\item \textit{Question-Answering task:} This task seeks to evaluate the summaries in term of their informativeness. 
This later is calculated using the number of correct answers which can be found in a summary for some questions generated from the source text.
Each automatic summary is compared manually to some answer keys for each input document, to decide if the answer is correct, partially correct or incorrect. 
ARS (\textit{Answer Recall Strict}) and ARL (\textit{Answer Recall Lenient}) metrics were defined to measure accuracy (see Equation \ref{eq:answer-recall}).
\begin{equation}
\label{eq:answer-recall}
ARS = \frac{n1}{n3}, 
ARL = \frac{n1 + (.5 * n2)}{n3}
\end{equation}
Where $ n1 $ and $ n2 $ are the numbers of correct and partially correct answers in the summary, and $ n3 $ is the number of questions answered in the key. 
\end{itemize}

\subsubsection{DUC/TAC}
\label{sssec:eval-duc}

In 2001, Document Understanding Conference\footnote{DUC: \url{http://duc.nist.gov/} [23 November 2016]} was launched as evaluation series in the area of text summarization. 
The aim of this workshop is to move forward the summarization research and enable researchers to test their methods in large-scale experiments. 

DUC 2004\footnote{Duc 2004 tasks: \url{http://duc.nist.gov/duc2004/tasks.html} [07 February 2017]} knew 5 evaluation tasks:
\begin{itemize}
\item \textit{Task 1 - Very short single-document summaries}: 
For each English document out of 50, a very short summary must be generated ($ <= $ 75 Bytes).

\item \textit{Task 2- Short multi-document summaries focused by TDT events}: 
For each set of English documents out of 50 sets, a short summary must be generated ($ <= $ 665 Bytes).

\item \textit{Task 3 - Very short cross-lingual single-document summaries}: 
For each English translation (automatic and manual) of 25 Arabic documents, a very short summary must be generated ($ <= $ 75 Bytes).

\item \textit{Task 4 - Short cross-lingual multi-document summaries focused by TDT events}: 
For each English translation (automatic and manual) of 25 Arabic document sets, a short summary must be generated ($ <= $ 665 Bytes).

\item \textit{Task 5 - Short summaries focused by questions}: 
For each set out of 50 English documents sets, a short summary must be generated ($ <= $ 665 Bytes) to answer the question in form "\textit{Who is X?}", where X is a name of a person or a group of persons.
\end{itemize}
The summaries which exceed the limit size are truncated, and no bonus is attributed to the summaries shorter than this.
The evaluation of tasks 1 to 4 uses ROUGE as a metric (ROUGE-1, ROUGE-2, ROUGE-3, ROUGE-4, and ROUGE-L).
In task 5, the summaries are evaluated in term of quality and coverage using ``\textit{Summary Evaluation Environment}" (SEE)\footnote{SEE: \url{http://www.isi.edu/licensed-sw/see/} [23 November 2016]}.
As for the pertinence to the question ``\textit{Who is X?}", human evaluators have been used.

DUC 2007 used AQUAINT corpus, which contains news articles from \textit{Associated Press}, \textit{New York Times} (1998-2000) and \textit{Xinhua News Agency} (1996-2000).
There have been two tasks: principal task and update task.
\begin{itemize}
\item \textit{Principal task:} For each topic of 25 documents, the contestants must generate a 200 words summary to answer one or more questions.

\item \textit{Update task:} The goal is to produce a 100 words multi-document summary as update, supposing that the user has already read the previous articles.
In this task, there are three clusters: cluster A with 10 documents for which the generated summary is not an update, cluster B with 8 documents for which the summaries must assume the user has already read those of cluster A, and cluster C with 7 documents which are more recent than those of cluster B.

\end{itemize}
The principal task is evaluated using many criteria:
\begin{itemize}
\item Linguistic form of each summary is evaluated manually using some criteria: grammar, non redundancy, references clarity, focus, structure and coherence.
For each criterion, the evaluator must give a score between 1 (not good) and 5 (very good).

\item The pertinence to a given topic is evaluated manually; For each summary, the evaluator must give a score between 1 (not good) and 5 (very good).

\item Automatic evaluation is used too, ROUGE-2, ROUGE-SU4 and BE.
\end{itemize}
As for update task, each summary is evaluated automatically using ROUGE-2, ROUGE-SU4, BE, and Pyramid.

Since year 2008, DUC has been included into the TAC conference as ``summarization" track.
The aim of this track is to develop ATS systems that afford short, coherent summaries of document.
This task is meant to promote a deep linguistic analysis for ATS.
It contains two tasks: the former DUC's ``update task" \citep{08-dang-owczarzak}, and a new one called ``opinion summarization task".
In the opinion summarization task, each system must generate well-organized, fluent summaries of opinions about specified targets, as found in a set of blog documents.
The questions are not simple, hence the answer can not be a named entity.
The evaluation is conducted manually using a nugget Pyramid created during the evaluation of submissions to the QA task.

In 2010 TAC's summarization track, a new task called "Guided summarization" replaced ``update summarization" one.
In this task, each system has to generate a 100 words summary from 10 news articles for each topic, where the topic belongs to a predefined category.
There are 5 categories: accidents and natural catastrophes, crises, health and safety, endangered resources, investigations and trials.
For a given topic, the contestants have to generate 2 summaries (for 2 sets: A and B):
\begin{itemize}
\item One for the set A, which is guided by a request.
\item The second (for set B) is the same as set A, but the summary must take in consideration that the user has already seen the documents of set A.
\end{itemize}
Each category has some aspects which have to be covered by the summary (for example, WHAT? WHY? WHEN? WHERE?).
To evaluate the content of summaries, Pyramid method is used. 
Readability and global sensibility are evaluated manually giving a score between 1 (not good) and 5 (very good).

\subsubsection{NTCIR}

The first ``\textit{NII Testbeds and community for Information access Research}" (NTCIR)\footnote{NTCIR: \url{http://ntcir.nii.ac.jp} [23 November 2016]} workshop was held in Tokyo, 1999. 
It was, originally, designed to enhance research in Japanese text retrieval. 
The second edition (2000-2001) had a ``Text summarization" task, which aims to collect data for text summarization and evaluate ATS systems.
The data was gathered from newspapers articles which were summarized by hand, to be used for research purposes. 
Two types of summaries were produced: extractive summaries (which are the important sentences in the text) and abstractive summaries.
Two tasks were proposed: intrinsic evaluation which contains two subtasks (extractive and abstractive) and extrinsic evaluation.
The process of each task is as follows \citep{01-fukusima-okumura}:
\begin{itemize}
\item \textit{Task A-1}: The aim is to extract pertinent sentences, where the number of the extracted sentences is  10\%, 30\%, 50\% of the original texts.

\item \textit{Task A-2}: The aim is to generate simple abstractive summaries.
The generated summaries must have a number of characters of 20\% et 40\% comparing to the original text.

\item \textit{Task B}: In this task, the summaries are produced based on some requests. 
For each request, the system has to search for one relevant document and use it to produce a summary. 
The length of the summary is not limited, but it has to be simple.
\end{itemize}
As to evaluate each task, the following metrics are used:
\begin{itemize}
\item \textit{Task A-1}: For each summary, the correct sentences are selected. 
Then, the metrics: recall, precision and F1 score are calculated using the number of sentences as a unit.
The final score of each system is the average of all summaries scores. 

\item \textit{Task A-2}: Two ways are used, an evaluation based on the content and a subjective evaluation.
In the first one, the distance between the two terms frequencies vectors representing the system's summary and the human summary is considered as the score.
In the second one, human evaluators are asked to evaluate the summaries based on two criteria: coverage and readability, and give a score of 1 (very good) to 4 (very bad).
Each one of them is given 4 summaries: 2 human summaries, the system's summary and a summary produced using LEAD method.

\item \textit{Task B}: In this task, human evaluators are given the requests and the generated summaries.
For each summary, they have to judge if it is relevant or not to the request.
Recall, precision and F1 score measures are calculated for each system based on the number of pertinent summaries.
One other measure is the time taken for each system to complete this task.
\end{itemize}

\subsubsection{MultiLing}

The Multiling workshop began as a task of TAC conference in 2011, which aims to evaluate language-independent summarization systems on many languages \citep{11-giannakopoulos-al}.
In this task, at least two languages out of seven must be processed by participant systems: Arabic, Czech, English, French, Greek, Hebrew and Hindi.
For each, it has to generate a summary of 240 to 250 words.
To create the test corpus, 10 topics were selected where every topic contains 10 news articles from Wikinews.
Then, these articles were translated to the other languages sentence by sentence.
To evaluate the generated summaries, the two types of evaluation has been used:
\begin{itemize}
\item Automatic evaluation: it aims to calculate the performance of the systems using some model summaries created by fluent speakers of each language.
Three methods have been used: ROUGE (ROUGE-1, ROUGE-2, ROUGE-SU4), MeMoG and AutoSummENG.

\item Manual evaluation: 
Overall responsiveness of a text was used, where each summary was given a score of 1 to 5 based on the content and the quality of the language.
When it covers all the important aspects of the original text and remain fluent, it will be attributed the score 5.
If it is unreadable, nonsensical or containing just trivial information, it will be attributed the score 1.
\end{itemize}
The topline system uses the model summaries (thus cheating) to select relevant sentences as summaries from original texts. 
The baseline system uses the centroid, extracted from a bag-of-words of same topic documents, to extract sentences using cosine similarity. 

In 2013, MultiLing went from a simple task of TAC to a workshop, which aims to test and promote multilingual summarization methods.
There were three tasks: ``Multi-document multilingual summarization" (MMS) \citep{13-giannakopoulos},  ``Multilingual single document summarization" (MSS) \citep{13-kubina-all} and ``Multilingual summary evaluation".
The 7 past languages used in MMS were used again along with three new languages: Chinese,  Romanian and Spanish.
The test corpus contains 10 topics for French, Chinese and Hindi, and 15 topics for the remaining languages. 
The evaluation methodology was the same as the 2011's, plus two automatic metrics: ROUGE-3 and NPowER.
%
The single document task introduces 40 languages, with a corpus of 30 documents for each language, created from wikipedia's featured articles.
To evaluate the summaries, automatic methods are used: ROUGE-1, ROUGE-2 and MeMoG.

Multiling 2015 had two more tasks: ``Call Center Conversation Summarization" (CCCS) and ``Online Forum Summarization" (OnForumS).
In CCCS task \citep{15-favre-al}, every system must generate abstractive summaries from call center conversations between a caller and an agent. 
The summaries must contain the caller's problem and the solution afforded by the agent. 
A corpus of French and Italian conversations was used, along with English translations of these two which makes them 3 languages. 
The two submitted systems had a hard time beating the three proposed baselines using ROUGE-2 as a metric. 
OnForumS task \citep{15-kabadjov-al} seeks to bring automatic summarization, argumentation mining and sentiment analysis all together. 
The data is a collection of English and Italian news articles along with the corresponding top 50 comments.  
Each system must generate some links between the article sentences and the comment sentences. 
Four research group submitted their systems which have been evaluated using crowd-sourcing which is evaluating the linked sentences by a human judge based on relation, agreement and sentiment between them.

\subsubsection{TREC}

Text retrieval conference\footnote{TREC: \url{http://trec.nist.gov} [23 November 2016]} is a metrics-based evaluation of TIPSTER Text program, which started in 1992. 
Its purpose is to provide the necessary infrastructure for large-scale evaluation, which can support research within the information retrieval community.

``\textit{Temporal Summarization}" is a task of TREC which started in 2013 and took place over 3 years.
The goal of this track is to develop systems for efficiently monitoring the information associated with a news event such as protests, accidents or natural disasters over time. 
The track has the following four main aims \citep{13-aslam-al}:
\begin{itemize}
	\item Developing low latency algorithms to detect sub-events,
	\item Modeling information reliability with a dynamic corpus,
	\item Understanding and addressing the sensitivity of text summarization algorithms in an on-line, sequential setting, and
	\item Understanding and addressing the sensitivity of information extraction algorithms in dynamic settings.
\end{itemize}
The track includes two tasks:
\begin{itemize}
	\item \textit{Sequential Update Summarization}: A system should emit relevant and novel sentences to an
	event. 
	A simulator is given as arguments the system to be evaluated, a time-ordered corpus, an event keyword query, the event start time and its end time.
	For each document in the time-ordered corpus, the system may choose the sentences relevant to the keyword query and novel comparing to the earlier timestamps.
	
	\item \textit{Value Tracking}: A system should emit accurate attribute value estimates for an event. 
	A simulator is given as arguments the system to be evaluated, a time-ordered corpus, an event keyword query, the event start time, the event end time and the event attribute. 
	First, the system is initialized with the query and the attribute to generate an initial estimated value. 
	Then, with every document, which is in the time-line, the system generate a new value along with the supporting sentence's ID if there is a change.
	
\end{itemize}
To evaluate the tasks, some metrics have been proposed (for detailed description and formulas, see \citep{13-aslam-al}):
\begin{itemize}
	\item The novelty and the relevancy of the updated summary to the event topic.
	The metric used to measure this is called the (normalized) Expected Gain metric
	($nEG(S)$).
	\item The coverage of the essential information for the event by the summary. 
	The metric used to measure this is called Comprehensiveness metric ($ C(S) $).
	
	\item The degree to which the information contained within the updates is outdated.
	This is measured by the Expected Latency metric ($ E[Latency] $).
	
\end{itemize}

Since 2016, The ``temporal summarization" track was merged with ``microblog track" to form a new track called ``\textit{Real-Time Summarization}" track.
RTS track\footnote{RTS: \url{http://trecrts.github.io} [23 November 2016]} is meant to explore novel and evolving information needed by users in streams of social media posts such as Twitter.
To achieve this goal, the track includes two scenarios:
\begin{itemize}
	\item \textit{Push notifications}: 
	In this scenario (Scenario A), the system must send relevant posts to the user's mobile phone as soon as they are identified. 
	These posts must be relevant to the user's interest, in time and novel.
	
	\item \textit{Email digest}:
	In this scenario (Scenario B), the system must identify a batch of 100 tweets according to a specific topic in a daily frequency.
	The summaries must be relevant to the user's interest and novel.
	The results are uploaded to NIST server after the end of the evaluation.
\end{itemize}
The evaluation takes 10 days, where all systems must listen to the Twitter sample stream using Twitter streaming API. 
A list of interest profiles will be provided to systems.
To evaluate the systems performances, two approaches are used: ``\textit{Live user-in-the-loop assessments} which is used only for the first scenario, and ``\textit{Post Hoc Batch Evaluations}" used for both scenarios.
\begin{itemize}
	\item \textit{Live user-in-the-loop assessments}: 
	Each system must push a notification to the TREC RTS evaluation broker as soon as it identifies relevant content to the users interest. 
	These notifications are immediately delivered to the mobile phones of a group of assessors who judge them using interleaved approach \citep{16-qian-al}.
	The assessor can judge the tweets in-time or later and send back the results to the evaluation broker using a mobile application.
	Each tweet is attributed a mention as relevant, relevant but redundant or irrelevant.
	
	\item \textit{Post Hoc Batch Evaluations}:
	A common pool is constructed based on the scenario's submissions where the depth of the pool is determined using the number of submissions and available resources.
	Tweets are assessed based on relevance: not relevant (get a gain of 0.0), relevant (get a gain of 0.5) or highly relevant (get a gain of 1.0).
	Then, relevant tweets are semantically clustered into groups with substantively similar information, the same as TREC 2015 Microblog evaluation.
	For the first scenario, EG-1, EG-0, nCG-1, nCG-0 and GMP scores are used, besides the latency which is the mean and the median difference between the time the tweet was pushed	and the first tweet in the semantic cluster to which the tweet belongs. 
	For the second scenario, nDCG@10-1 and nDCG@10-0 are used.
	
\end{itemize}

\section{Challenges}
\label{sec:challenges}

There exists a lot of challenges which harden the advance of ATS. 
One of these problems is the absence of a precise definition of what should be included in a summary. 
Over time, several works were conducted to generate the perfect summary which must be informative in one hand, and not redundant in the other. 
The first challenge is the definition of a good summary, or more precisely how can a summary be generated. 
Our needs for a summary are good indicators of what it should be: extractive or abstractive, generic or query-driven, etc. 
Even if we get the idea how humans usually summarize, implementing it will not be easy. 
Designing a powerful automatic text summarizer needs a lot of resources either tools or corpora. 
Another type of challenges is the summary informativeness; How can a system imitate human beings in summarizing task?
The coherence of the summary is one of the challenges that stood for years. 
Mostly, ATS methods aim to generate an informative summary, but when it comes to the summary's readability it is a matter of future work.  

\subsection{Definition}

A good summary can be defined as the shortest and more informative grammatically correct one without redundancy. 
This definition seems simple, but when we want to implement such idea, we will come across a lot of choices. 
A text, usually, contains a main topic and some satellite topics. 
For instance, if we talk about PCs, we may also include some commercial notions to express the market shares.
The choice of what topic to favor raises the question: what is a good summary?
If the summarization system is query-based, it is settled that the most similar the topics are to the request the more they are relevant. 
Nevertheless, Some may argue that adding relevancy to the main topic along with the request can be helpful for the user.

When it comes to generic summarization systems, it is more difficult to decide what is better for the reader.
Some researches try to capture the most relevant sentences from each topic to form a summary (eg. \citep{11-song-al}). 
The summary can cover all the topics in the text, but sometimes it got far from the point.
Others take only the most relevant sentences to the main topic and ignore the secondary topics. 
This leads to a summary more focusing on the main topic, and ignoring the others which can be as important as the main topic.
Sure we have to give more importance to the main topics, but the satellite ones count too. 
This is what some researches are trying to achieve; a balance between the main topic and other topics that may contain some useful information (eg. \citep{12-zhang-al,13-aries-al}). 

\subsection{Summary informativeness}

The main goal of a summary is to afford a representative text of the original document. 
It must cover the essential information using few words. 
Which means, the summary must retain more information with less redundancy. 

\subsubsection{Summary coverage}

A summary must cover the most important content of the original document. 
When we read a summary, we must get an idea on what is its original document about, or what the document can tell about a specific request. 
Information quantity has been the main focus of intrinsic evaluation methods, and evaluation workshops as well. 
A lot of advance has been made, so it can be considered as a minor problem.
For example, in MultiLing 2015, the systems had recall scores close to the top-line system which is a system that cheats in order to generate the summaries.

\subsubsection{Summary conciseness}

A good summary must not contain redundant information, leaving the space for more useful information to be included. 
Most systems reorder sentences using their scores, so they can highly score two similar sentences and include them to the summary, especially in case of multi-document summarization.
When generating summaries, a system must exclude similar sentences even if they score more than others.
A very known work which takes redundancy in consideration is MMR (Maximal marginal relevance) \citep{98-carbonell-goldstein}.
The authors score each candidate sentence $ s_i $ of a document $ D $ (which is not already included to the summary $ S $) using two scoring similarities: $ \text{\textit{sim}}_1 $ which calculates the similarity of $ s_i $ to a query $ Q $, and $ \text{\textit{sim}}_2 $ which calculates the similarity of $ s_i $ to another sentence $ s_j \in S $.
The idea is to maximize MMR score given in Equation \ref{eq:mmr}.
\begin{equation}
\label{eq:mmr}
MMR = \arg \max \limits_{s_i \in D\backslash S} [ \lambda\, \text{\textit{sim}}_1 (s_i, Q) - (1-\lambda) \max\limits_{s_j \in S} \text{\textit{sim}}_2 (s_i, s_j)]
\end{equation}
Where, $ \lambda $ is a parameter that balances between the relevance ($ \lambda = 1 $) and the diversity ($ \lambda = 0 $).
The authors use cosine similarity to calculate $ \text{\textit{sim}}_1 $ and $ \text{\textit{sim}}_2 $.

More complex methods are proposed to find a trade-off between the coverage and the conciseness of the generated summary.
For instance, \citet{12-nishikawa-al} model text summarization as a knapsack problem to generate a summary with a maximum coverage, selecting sentences with as many information units (unigrams and bigrams) as possible.
Then, they add a constraint to prevent adding a unit that already exists in the summary.

The two previous methods try to score the sentences based on their coverage and conciseness in the same time.
A more simpler method is to score the sentences using just the coverage then control redundancy while selecting summary ones.
In \citep{15-aries-al}, cosine similarity and a threshold are used to decide two sentences similarity. 
The most scored sentence is selected into the summary, and each time the next one to be added is tested whether it is similar to the last added one, if not it will be included to the summary.

Redundancy is not just a property of texts, it can be found even in videos. 
\citet{16-bhaumik-al} divide a video into multiple shots from which key-frames are extracted. 
Then, they remove intra and inter-shots redundant frames to form the final video summary.

\subsection{Summary readability}

Extractive summarization methods are based on extracting pertinent units (sentences often) from the source document(s). 
The extracted units are used to form the summary, which can contain dangling anaphora references and sentences which are ordered badly.
Abstractive methods seem to handle the anaphora problem and the order of sentences, since the generation is from a semantic representation. 
But the problem is, can the generator afford a good grammar with a fluent language? 
Summary readability has been in ``future works" section for years, a future that has never come. 

\subsubsection{Reference clarity}

When we extract summaries, some references remain unsolved; such as personal pronouns. 
That way, the reader can't find out who or what is involved in the sentence.
So, to generate a good summary we have to replace the first mention of a reference in the summary. 
In \citep{16-bayomi-al}, it is found that the introduction of anaphoric resolution can slightly improve the readability of the summary. 
Also, the authors confirm that the impact of anaphora resolution (AR) varies from one domain to another which means it improves some domains summaries more than others.

The resulted summary's readability can be enhanced by resolving anaphora, but it can even improve the process of summarization itself. 
\citet{07-orasan-stafford} try to test the effect of AR on the summarization process using the TF*IDF for scoring and three different AR methods.
The informativeness of generated summaries improves when AR is used before summarization. 
In the other hand, it has no correlation with the performance of the anaphora resolver used to improve
the frequency counts. 
\citet{07-steinberger-al} incorporate AR into a more complex scoring method (LSA representation \citep{01-gong-liu}) to improve its performance even with an imperfect anaphoric resolver.
To improve the coherence, they propose an algorithm to replace incorrect anaphoric relations in the summary by their respective original expressions in the text, which leads to a precision of 69\%.

\subsubsection{Coherence}

Most summarization methods are based on extracting relevant sentences and present them as they are. 
It would be a good idea if these sentences are reordered since they are extracted from different parts of a text. 
When reading the summary, the reader has to feel the flow of ideas without just jumping from idea to another.

In single document summarization, it seems right to present them as ordered in the original text. 
That is, reordering them using their original positions \citep{99-mckeown-al,00-radev-al,02-lin-hovy}. 
But, sometimes it will be better to move a sentence towards another due to their similarities. 
Methods which use rhetorical relations, such as \citep{98-marcu}, do not suffer this problem since they keep the relations between sentences.

In multi-document summarization, sentence reordering is important since the sentences come from different sources. 
In this type of summarization, the main topic is the same, but the secondary topics are not always the same in each document.
Moreover, sentences in each source document are not ordered in the same flow of ideas. 
Some works tried to solve this problem differently. 
One solution is ``Majority Ordering" \citep{02-barzilay-al}, where similar sentences are grouped together into themes. 
Then, a graph representing the local order is created, where each node represents a theme. 
Two nodes are connected with an arc if, in a document, one sentence of the first node's theme is followed by another of the second node's theme.
The weight of this arc is the number of documents where this order is observed.
To infer the global order of the graph, an approximation algorithm is used.
Other variations of this method can be found in the works of \citet{05-bollegala-al}; \citet{08-ji-nie}.


\subsection{Resources}

One of the most problematic challenges in ATS is the lack of resources. 
Nowadays, there are many powerful tools for stemming, parsing, etc. compared to the past. 
Nevertheless, there is the challenge of finding the adequate ones for a certain problem of summarization. 
Besides, annotated corpus for ATS can be seen as a challenge as well.
In this section, we will try to present some freely available resources which can be helpful to researchers.

\subsubsection{Toolkits}

Generally, when we create a new method, we have to compare it to other existing methods.
In summarization, we can generate summaries with the different methods including ours and evaluate them using the earlier presented evaluation metrics.
A well known summarizer is MEAD\footnote{MEAD: \url{http://www.summarization.com/mead/} [21 January 2017]}, which includes baseline summarization algorithms besides others.
%
Semantic\_summ\footnote{semantic\_summ: \url{https://github.com/summarization/semantic_summ} [21 January 2017]} \citep{15-liu-al} is another system which uses semantic representations to generate abstractive summaries.
As example of neural networks in ATS, we can cite NAMAS\footnote{NAMAS: \url{https://github.com/facebook/NAMAS} [21 January 2017]} \citep{15-rush-al}.
AllSummarizer\footnote{AllSummarizer: \url{https://github.com/kariminf/allsummarizer} [21 January 2017]} \citep{15-aries-al} is a statistical extractive summarizer designed for multilingual summarization.
ABSTAT\footnote{ABSTAT: \url{https://github.com/siti-disco-unimib/abstat} [21 January 2017]} \citep{15-palmonari-al} produces a summary of linked data sets which make use of ontologies to describe the semantics of their data.
Berkeley-doc-summarizer\footnote{Berkeley-doc-summarizer: \url{https://github.com/gregdurrett/berkeley-doc-summarizer} [21 January 2017]} \citep{16-durrett-al} uses machine learning, compression and anaphora resolution to generate single document summaries.
This list comprises a little sample of the wide ATS systems, which proves the emergence of openness support among ATS systems.

Summarization methods can be hard to be implemented, especially when they are heavily based on complex NLP functionality. 
NLP methods are more advancing by time, such as tokenizers, stemmers, syntactic parsers, etc.
Unfortunately, many original systems used to test these methods are not available either openly or commercially. 
There are some languages that lack basic NLP tools such as tokenizers and stemmers.
Recently, there seems to be a more interest on affording open source toolkits to support research among NLP community.

\subsubsection{Corpora} 

Many research methods use machine learning to train the system on a labeled corpus and then test it on another one. 
The problem, in many cases, is the absence of adequate corpora for ATS. 
Text summarization task is affected by the genre of training corpus (news, novels, etc.), which imply the creation of diverse corpora based on the purpose of the system.
Some researches tried to create methods with semi-supervised learning to solve this problem \citep{02-amini-gallinari}.
To adjust this problem, some workshops like MultiLing propose tasks for summarization corpora creation. 
Also, they afford training and evaluation corpora for automatic text summarization.

\subsection{Evaluation}

Comparing a generated summary automatically to some man-made summaries is good, but not enough. 
The summarization system still can generate a good summary which is not even close to the reference summaries. 
Also, using reference summaries is more likely to work with extractive systems better than abstractive ones. 
This is a brief summary of some challenges:
\begin{itemize}
	\item People write summaries differently, therefore we can not say this is the best summary for sure. 
	Hiring professionals to evaluate automatic summaries can be very expensive in term of coast and time. 
	Also, a summary can be judged differently from a person to another due to human variation.
	
	\item To face the problems with human evaluation, especially the evaluation time, automatic methods can be used. 
	Most automatic methods evaluate just the content without the quality aspect. 
	Also, they are more adequate for extractive summarization methods than the abstractive ones. 
	Using words or n-grams in a reference summary and comparing them to those of the automatic summary can favorize certain systems over other better ones. 
	
	\item Automating quality evaluation such as grammar, reference clarity and coherence can be very challenging. 
	The challenge is the same as in ATS one, discussed earlier.
	This needs more profound linguistic analysis of the summary, with accurate and powerful tools.
	Also, the methods that can detect such flaws can be used to fix them in the first place.
	For example, if we can detect automatically that a sentence must come earlier than another, this can be added to the summarization systems to fix the problem.
	
\end{itemize}

\section{Conclusion}
\label{sec:conclusion}

Automatic text summarization continues to gain more importance due to large amount of information nowadays. 
In this article, we reviewed different ATS classifications which can beneficial to know in order to create new methods. 
Then, we discussed some approaches used to generate the final summary. 
After presenting the summarization techniques, we must present the evaluation part which is a domain in its own. 
There are different evaluation methods, which estimate the capacity of a system to imitate human summarization task.
These methods have been used in different workshops intended for ATS systems evaluation. 
We tried to capture and discuss some of the challenges which can be the definition of the summary itself, informativeness, readability, availability of resources and evaluation.

In general, before summarizing we must define what is the summary for. 
By classifying our intended method (is it request-based? is it extractive or abstractive?, etc.), we can define what could be a good summary in our case. 
A good summarization method for a specific genre like news could be of no use for novels, etc. 
As for the approach, many factors can define if we use statistical, linguistic, centroid, etc. 
Statistical methods are more fast and simple to be designed, since they are based on simple techniques such as TF-IDF. 
To be more accurate, deep linguistic methods have to be used; for example, TF-IDF can be boosted by introducing words synonyms.
Linguistic methods can capture more aspects in the original text, such as rhetorical relations, semantic relations, etc., but it can be difficult to implement such methods without adequate resources.
The absence of resources, either tools or corpora, is a challenging problem not just for linguistic methods but also for statistical ones.
Machine learning is often used with both statistical and linguistic approaches to train the system using a labeled corpus. 
Evaluating different methods is challenging too, since a summary has to be evaluated using many aspects. 
Informativeness is mostly covered by the different workshops, but what is mostly ignored is the summaries readability.
It is an aspect to which recent methods may pay more attention.

\bibliographystyle{unsrtnat}
{\footnotesize
\bibliography{ATS}
}

\end{document}